\documentclass[11pt]{article}
\usepackage[preprint]{acl} 

\usepackage{times}
\usepackage{latexsym}
\usepackage[T1]{fontenc}
\usepackage[utf8]{inputenc}
\usepackage{microtype}
\usepackage{inconsolata}
\usepackage{graphicx}

\usepackage{amsmath,amssymb}
\usepackage{booktabs}
\usepackage{multirow}
\usepackage{tabularx}
\usepackage{array}
\usepackage{url}
\usepackage{xcolor}

\newif\ifupdate\updatefalse

\newcommand{\method}{\textsc{AutoVerifier}}
\newcommand{\correct}{\textsc{Correct}}
\newcommand{\incorrect}{\textsc{Incorrect}}
\newcommand{\abstain}{\textsc{Abstain}}

\newcommand{\Pfb}{\ensuremath{P_{\mathrm{fb}}}}
\newcommand{\Vfb}{\ensuremath{V_{\mathrm{fb}}}}
\newcommand{\RUL}{\mathrm{RUL}}
\newenvironment{avcompactitemize}{%
  \begin{list}{$\bullet$}{%
    \setlength{\leftmargin}{1.4em}%
    \setlength{\itemsep}{2pt}%
    \setlength{\parsep}{0pt}%
    \setlength{\topsep}{0pt}%
    \setlength{\partopsep}{0pt}%
  }%
}{%
  \end{list}%
}
\newlength{\avmaintablewidth}
\newcommand{\avtabgroup}[1]{%
  \specialrule{0.35pt}{0pt}{0pt}%
  \multicolumn{6}{@{}l@{}}{%
    \begingroup
    \setlength{\fboxsep}{0pt}%
    \colorbox{gray!10}{\makebox[\avmaintablewidth][l]{\rule[-0.45ex]{0pt}{2.7ex}\hspace{4pt}\textit{#1}}}%
    \endgroup
  }\\[-0.12ex]\specialrule{0.35pt}{0pt}{0.25ex}%
}
\newlength{\avboxwidth}
\newcommand{\avcasebox}[2]{%
  \par\smallskip\noindent
  \begingroup
  \setlength{\fboxsep}{0pt}%
  \setlength{\fboxrule}{0.5pt}%
  \setlength{\avboxwidth}{\dimexpr\linewidth-2\fboxrule\relax}%
  \fcolorbox{gray!55}{gray!5}{%
    \begin{minipage}{\avboxwidth}
    \colorbox{gray!18}{\parbox{\avboxwidth}{\strut\hspace{6pt}\bfseries\small #1\strut}}%
    \par\vspace{5pt}%
    \hspace{6pt}\begin{minipage}{\dimexpr\avboxwidth-12pt\relax}
    \small #2
    \end{minipage}\par\vspace{6pt}
    \end{minipage}%
  }%
  \endgroup
  \par\smallskip
}

\title{AutoVerifier: Residual-Guided Non-Parametric Optimization for Reference-Based Answer Verification}

\author{
  \textbf{Zebei Zhao\textsuperscript{1}},
  \textbf{Zhihao Shi\textsuperscript{1}},
  \textbf{Minqi Shi\textsuperscript{2}} \\
  \textsuperscript{1}University of Science and Technology of China \\
  \textsuperscript{2}Beihang University \\
  \texttt{\{zbzhao,zhihaoshi\}@mail.ustc.edu.cn} \\
  \texttt{Sherrie@buaa.edu.cn} \\
}

\begin{document}
\maketitle

\begin{abstract}
Reference-based verifiers are important for evaluating reasoning models and providing accurate outcome rewards in reinforcement learning with verifiable rewards. To improve verification accuracy, prior work has explored rule-based, model-based, and tool-augmented verifiers for checking answer equivalence across diverse answer forms. However, the equivalence of answer forms such as $1+3.14$ and $1+\pi$ may depend on the question and scoring criterion. We frame such implicit assumptions as verifier inductive biases. To address this challenge, we propose \method{}, a residual-guided non-parametric optimization method that learns these biases from recurring verifier errors. Specifically, \method{} records these biases in rule cards and promotes them to code modules or prompt guidance only after replay validation detects no direct regressions, keeping accepted updates auditable, editable, and reusable. Experiments on four verifier benchmarks demonstrate that \method{} outperforms state-of-the-art verifiers by a large margin.
\end{abstract}

\section{Introduction}
\label{sec:intro}

Answer verification is widely used to check outputs of reasoning models in benchmark evaluation \citep{yan2025verifybench,chen2025xverify,liu2025compassverifier} and to provide outcome rewards for reinforcement learning with verifiable rewards \citep{shao2024deepseekmath,guo2025deepseekr1,yu2025dapo}. In reference-based verification, a verifier takes a question, a reference answer, and a candidate response as input to determine the correctness of the candidate response. As reasoning models produce longer and more diverse responses, exact matching alone is insufficient; the verifier must judge answer equivalence across diverse answer forms.

\begin{figure}[t]
\centering
\includegraphics[width=\linewidth,height=0.557\linewidth]{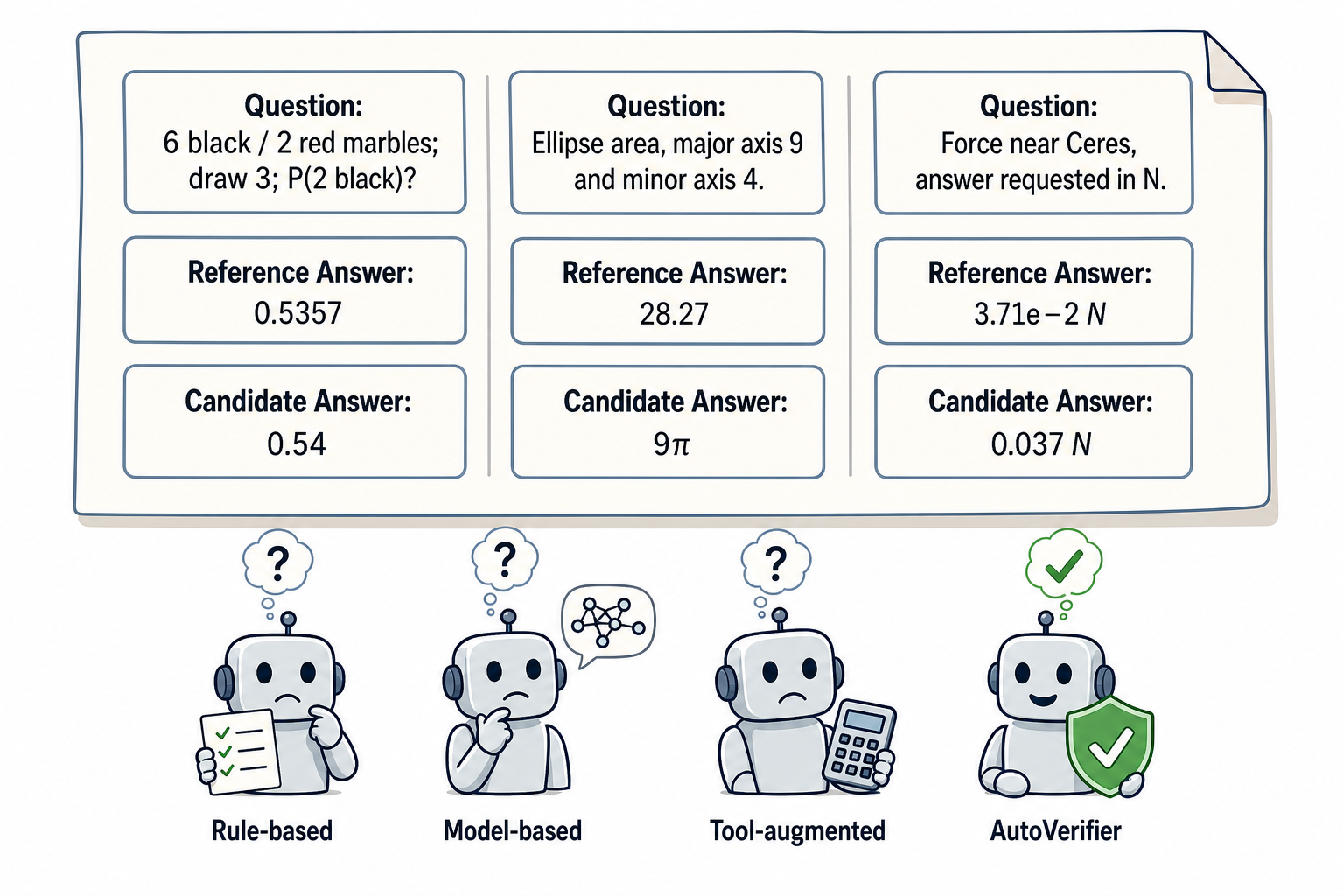}
\caption{Answer equivalence can depend on implicit assumptions. The examples cover rounded probabilities, symbolic forms for rounded numeric references, and scientific notation with units. Whether each pair is equivalent depends on the question and scoring criterion.}
\label{fig:scale_problem}
\end{figure}

Recent verifier work has improved either answer matching coverage or made parts of comparison logic more explicit. Rule-based verifiers and libraries such as Math-Verify make comparison logic inspectable as code for fixed checks such as symbolic equality and numerical tolerances \citep{kydlicek2024mathverify}. Model-based verifiers and judge models improve coverage for challenging answer matching cases such as formulas, sequences, and multi-part answers, with recent reward models using generated rationales or critiques to support judgment \citep{chen2025xverify,liu2025compassverifier,chen2025judgelrm,liu2025grm,xia2025agentrm}. Tool-augmented verifiers add executable computation for scientific questions \citep{feng2025cosineverifier}.

Despite these advances, a key gap remains when human annotations reflect implicit assumptions about which candidate answers should count as equivalent for a given question and scoring criterion. Figure~\ref{fig:scale_problem} illustrates this issue with rounded probabilities, symbolic forms for rounded numeric references, and scientific notation with units. For example, $1+3.14$ is not equivalent to $1+\pi$ in an exact computation problem, but can be acceptable in an applied numerical problem where a rounded value of $\pi$ is expected. Following \citet{baxter2000model}, we frame these implicit assumptions as verifier inductive biases to be learned from recurring verifier errors. Rule-based and tool-augmented verifiers remain limited by fixed comparison logic. Recent model-based verifier work covers complex answer matching, but treats ambiguous numerical acceptability thresholds as outside the binary verification setting \citep{liu2025compassverifier}. Moreover, criteria learned by model-based verifiers remain embedded in model behavior rather than exposed as editable comparison logic. The unresolved problem is therefore to learn verifier inductive biases automatically from evidence while keeping accepted updates auditable, editable, and reusable.

To address this problem, we propose \method{}, a non-parametric optimization method that learns verifier inductive biases from recurring verifier errors. Specifically, \method{} groups recurring errors, records the corresponding biases in rule cards with supporting examples and counterexamples, and promotes a card only after evidence and replay validation confirm no direct regressions. Cards with deterministic comparison logic become code modules after replay validation, while cards that require model judgment are incorporated into prompt guidance for the model-based fallback.

Experiments on four benchmarks demonstrate that \method{} outperforms existing reference-based verifier baselines. \method{} reaches 93.05\% macro accuracy, exceeding the reported accuracy of rule-based, model-based, scientific, and tool-augmented verifiers. Compared with the prompt-only setting, adding code modules improves macro accuracy by +1.76 points, indicating that learned comparison logic contributes beyond prompt rewriting. The same direct decisions also reduce fallback calls by 32.13\%.

\noindent Our key contributions can be summarized as follows:
\begin{avcompactitemize}
    \item We identify implicit annotation assumptions as \textbf{verifier inductive biases} that can be learned from recurring verifier errors.
    \item We propose \textbf{\method{}}, a residual-guided non-parametric optimization method that records recurring verifier errors as rule cards and promotes them through replay validation.
    \item We conduct experiments on four reference-based verification benchmarks, demonstrating that \method{} improves over a prompt-only baseline, outperforms reported verifier baselines, and reduces fallback model calls.
\end{avcompactitemize}

\section{Related Work}
\label{sec:related}

\paragraph{Answer verifiers.}
Answer verifiers commonly rely on predefined rules, model judgments, or tool pipelines. Rule libraries such as Math-Verify encode extractors, parsers, normalizers, numeric tolerances, symbolic equivalence checks, and abstention conditions as inspectable procedures \citep{kydlicek2024mathverify}; scientific verifiers add domain logic for algebraic equivalence, unit conversion, and notation handling \citep{zheng2025sciverifier}. This auditability still requires manual coverage growth, and subtle equivalence remains failure prone \citep{huang2025pitfalls}. Model-based verifiers broaden coverage beyond fixed answer matching rules \citep{chen2025xverify,liu2025compassverifier,chen2025judgelrm,liu2025grm,xia2025agentrm}, while tool-augmented verifiers such as CoSineVerifier execute computation or unit conversion for scientific questions \citep{feng2025cosineverifier}. However, rule and tool pipelines still rely on fixed equivalence checks, while model-based acceptance criteria remain embedded in model behavior. \method{} targets the middle ground by recording recurring verifier errors in rule cards that become auditable code modules, while unresolved cases remain with the model-based fallback.

\paragraph{Non-parametric updates from experience.}
The learning beyond gradients perspective studies systems that improve by editing external state, including code, tests, memory, and evaluation records, rather than model parameters \citep{weng2026learning_beyond_gradients}. Procedural memory work such as Skill-Pro promotes reusable procedures through activation, execution, termination, and validation \citep{mi2026skillpro}; related agents and program search systems evolve skills, tools, scaffolds, or code from experience \citep{wang2023voyager,hao2026recreate,zheng2025skillweaver,romera2024funsearch,novikov2025alphaevolve}. \method{} instantiates this non-parametric update pattern for verification, where recurring verifier errors are recorded as rule cards and accepted cards become code modules or prompt guidance after replay validation.

\begin{figure*}[t]
\centering
\includegraphics[width=0.98\textwidth,height=0.47\textwidth]{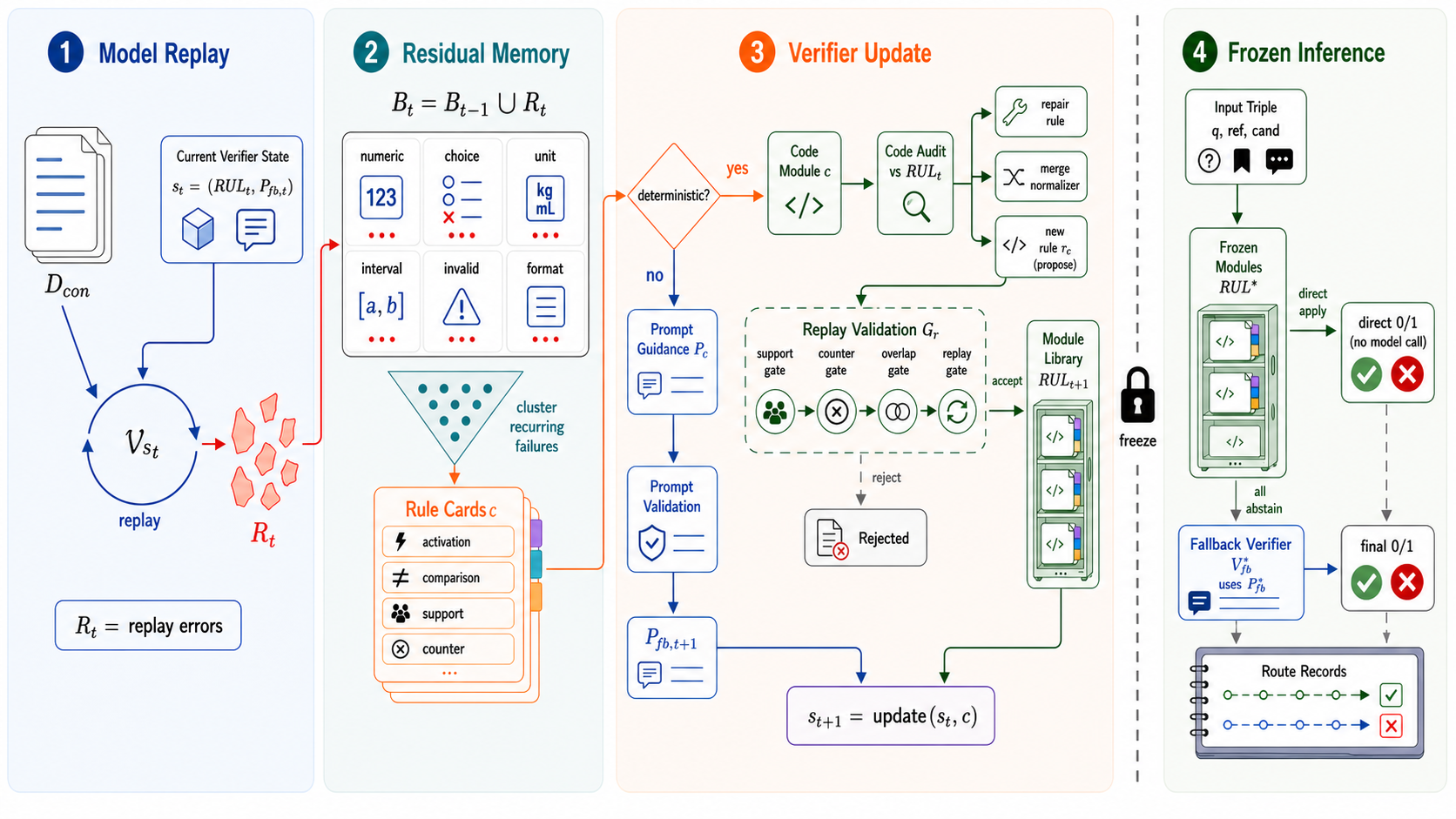}
\caption{Overview of \method{}. During construction, the current verifier is replayed on 5,000 examples, and recurring verifier errors are recorded as rule cards. Cards with deterministic comparison logic enter code audit and replay validation before becoming code modules, while cards that require model judgment are incorporated into prompt guidance. After freezing, code modules make direct decisions when they apply, and examples for which all code modules abstain go to the model-based fallback.}
\label{fig:system_overview}
\end{figure*}

\section{Method}
\label{sec:method}

\method{} learns verifier inductive biases from recurring verifier errors with a model-based fallback verifier. Construction replay exposes answer forms that the current verifier mishandles, and accumulated errors are grouped into recurring patterns before any update is proposed. A recurring pattern becomes a rule card that records the bias, its support examples, its counterexamples, and the conditions under which a code module should abstain. Here, a code module is a deterministic verifier unit that can return \correct{}, \incorrect{}, or \abstain{}. Cards with deterministic comparison logic enter code audit and replay validation before becoming code modules, while cards that require model judgment are incorporated into prompt guidance. This section defines the verification setting, the verifier update, the rule card paths, the replay validation procedure, and the frozen inference procedure.

\subsection{Problem Formulation}
\label{sec:setup}
We focus on pointwise reference-based answer verification. Each example is
\[
x_i = (q_i, a^\star_i, \hat a_i, m_i),
\]
where $q_i$ is a question, $a^\star_i$ is the reference answer, $\hat a_i$ is a candidate response, and $m_i$ contains optional metadata such as domain or answer type. The binary label $y_i \in \{0,1\}$ indicates correctness (\correct{} or \incorrect{}).  

The construction problem starts from a fixed base verifier $V_{\mathrm{base}}$ and updates the verifier around it. We denote the construction pool by $D_{\mathrm{con}}$. A residual is a construction example where the current verifier disagrees with $y_i$, and these residuals provide evidence for rule extraction. Support examples and counterexamples are construction examples used to validate a proposed rule card before benchmark evaluation. A recurring error pattern is treated as evidence that the current verifier lacks a verifier inductive bias for deciding whether a recurring answer transformation is equivalent or nonequivalent. Patterns that can be expressed as deterministic code become updates to a \textbf{module library} of code modules after code audit and replay validation, while patterns that require model judgment are handled by a \textbf{model-based fallback verifier} $\Vfb$ implemented by the fixed model with fallback prompt $\Pfb$.

\subsection{Verifier State Updates}

At construction round $t$, the verifier state is
$s_t=(\RUL_t,P_{\mathrm{fb},t})$, where $\RUL_t$ is the module library and $P_{\mathrm{fb},t}$ is the fallback prompt. Replaying this state on construction data yields the residual set:
\[
R_t = \{x_i \in D_{\mathrm{con}} : V_{s_t}(x_i) \neq y_i\}.
\]
The residual buffer accumulates recurring errors before rule card drafting. Let $V_{s_t}$ denote the verifier induced by $s_t$ under the fixed routing rule. We maintain a construction residual buffer $\mathcal{B}_t = \bigcup_{\tau \le t} R_\tau$. Here $R_t$ is the residual set for the round, while $\mathcal{B}_t$ stores accumulated residuals used for later clustering. The coordinator groups recurring error patterns in $\mathcal{B}_t$ by answer object, surface form, source domain, and observed error type before drafting rule cards.

Rule cards specify how a bias should be applied. Each card records the answer comparison procedure, activation region, abstention behavior, and support examples and counterexamples that validate the proposal. A card with deterministic comparison logic proposes a code update $r_c$, which may repair or merge an existing module path or add a new code module when the behavior is not already covered; a card that requires model judgment proposes prompt guidance $P_c=\mathrm{Guide}(P_{\mathrm{fb},t},c)$.

The promotion check serves as conservative validation because a candidate may expand direct decisions only if replay shows no direct regression on protected construction examples. Let $G_r(c;s_t)$ be the conjunction of support, counterexample, overlap, and full replay validation for a code module candidate, and let $G_p(c;s_t)$ be the validation check for prompt guidance. Let $\mathcal{C}_{\mathrm{code}}$ and $\mathcal{C}_{\mathrm{guide}}$ denote code and guidance cards. We write $s_t\oplus r_c$ for applying a code update that passes replay validation to $\RUL_t$, and $s_t\oplus P_c$ for revising $P_{\mathrm{fb},t}$ with guidance from card $c$. The state update is
\[
s_{t+1} =
\begin{cases}
s_t\oplus r_c, & c\in\mathcal{C}_{\mathrm{code}},\ G_r(c;s_t),\\
s_t\oplus P_c, & c\in\mathcal{C}_{\mathrm{guide}},\ G_p(c;s_t),\\
s_t, & \mathrm{otherwise.}
\end{cases}
\]
Thus, the construction loop uses a proposal followed by conservative validation, while updating verifier state around a fixed model rather than model parameters.

\paragraph{Replay validation objective.}
The replay validation objective formalizes conservative promotion as supported coverage under zero direct error constraints. Let $V_0=V_{\mathrm{base}}$.  In the first construction round, residuals are collected from the fixed base verifier:
\[
R_0 =
\{x_i:\ x_i\in D_{\mathrm{con}},
\ V_{\mathrm{base}}(x_i)\neq y_i\}.
\]
In later rounds, residuals are collected by replaying the current verifier state $(\RUL_t,P_{\mathrm{fb},t})$ on construction data.  Let $K_{\RUL_t}(x)\in\{0,1,\bot\}$ be the direct output of the current module library, where $\bot$ means that all activated code modules abstain.  For a candidate rule card, let $S^+$ denote positive support examples, $S^-$ denote counterexamples, and $C$ denote the larger construction replay pool. \method{} constructs code modules $\RUL$ to maximize supported direct coverage under replay constraints with no regressions:

\[
\mathrm{Cov}_{S^+}(\RUL) =
\frac{1}{|S^+|}\sum_{x_i\in S^+}
\mathbf{1}[K_{\RUL}(x_i)\neq \bot].
\]

\[
\begin{aligned}
\mathrm{Err}_A(\RUL) =
\sum_{x_i\in A} \mathbf{1}\big[
&K_{\RUL}(x_i)\neq \bot\\
&\wedge K_{\RUL}(x_i)\neq y_i
\big].
\end{aligned}
\]

\[
\begin{aligned}
\max_{\RUL}\quad & \mathrm{Cov}_{S^+}(\RUL) \\
\mathrm{s.t.}\quad
& \mathrm{Err}_{S^+\cup S^-}(\RUL)=0,\\
& \mathrm{Err}_C(\RUL)=0.
\end{aligned}
\]

Support examples encourage intended coverage, while counterexamples and replay examples require the code module to abstain or return the construction label on nearby negatives. In this sense, the promotion check serves as conservative validation because a candidate may add supported direct decisions but cannot introduce direct regressions on protected construction replay. The fallback verifier handles abstained or novel cases after the verifier is fixed.

\subsection{Rule Cards and Verifier Updates}

\method{} treats residuals as evidence of recurring verifier error patterns. Repeated residuals are summarized into structured profiles containing source, domain, answer object, candidate surface form, observed label mismatch, and short error hints. The coordinator first groups profiles by answer object and observed error type, then filters out groups defined only by benchmark identity or a single example. Only after this clustering step are surviving groups written as rule cards. Each rule card is a structured proposal with activation, comparison procedure, abstention rule, support examples, counterexamples, and replay evidence.

\begin{avcompactitemize}
    \item Rule cards with deterministic activation and comparison logic are compiled into \textbf{code modules}, which repair an existing module path or add a new direct decision path to the reusable library.
    \item Rule cards that cannot be encoded safely as deterministic code are incorporated into prompt guidance, which updates the textual prompt processed by the fixed model but never makes direct decisions.
\end{avcompactitemize}

This proposal and replay loop implements the objective above. Accepted code modules expand supported direct coverage while preserving the replay constraints.

\paragraph{Code module and prompt guidance paths.}

\method{} implements two update paths. A rule card with deterministic logic first enters code audit against the current module library. A missing guard or overbroad normalizer triggers repair or merge of the existing module path, while uncovered behavior triggers a new code module. Cards that require model judgment are incorporated into prompt guidance, preserving lessons useful for judgment but unsafe for direct deterministic decisions. A deterministic coordinator applies the checks above and stores rejected candidates for later rounds.

\subsection{Replay Validation for Promotion}

This subsection defines the replay validation used to promote a candidate code module. Each candidate code module must expose activation, execution, and termination structure, specialized to binary answer verification:

\begin{avcompactitemize}
    \item \textbf{Activation condition} $\alpha_r$: defines when the direct comparison may run.
    \item \textbf{Execution procedure} $\pi_r$: performs the deterministic answer comparison procedure.
    \item \textbf{Termination rule} $\tau_r$: returns \correct{}, \incorrect{}, or \abstain{}.
\end{avcompactitemize}

Candidate code modules undergo four construction data checks:
\begin{avcompactitemize}
    \item \textbf{Support:} target support examples must trigger the intended activation path and return the construction label.
    \item \textbf{Counterexamples:} nearby negative examples must abstain or return the construction label.
    \item \textbf{Overlap:} the candidate must add coverage rather than duplicate existing checks.
    \item \textbf{Replay:} full construction replay must not introduce new direct errors.
\end{avcompactitemize}

The checks enforce narrow activation, nearby counterexamples, and replay protected promotion. Accepted updates are compiled into deterministic code, while failed candidates are stored to avoid repeated proposals. Support examples come from recurring verifier errors or fixtures that exercise the intended activation path. Counterexamples are nearby construction cases that share surface form but differ in answer object, polarity, unit, interval endpoint, terminal slot, or invalid output status. Numeric tolerances, unit normalizers, terminal extraction windows, and closed vocabulary sets are written into the rule card during construction and can be changed only through construction replay. If a candidate changes a construction example that the prompt labels correctly into a direct error, duplicates an earlier check without verified gain, or depends on an unchecked semantic assumption, it is rejected.

Appendix~\ref{app:case_study} gives a concrete rule card example, including activation, execution, abstention, counterexamples, and promotion evidence.

\begin{table*}[!t]
\centering
\footnotesize
\renewcommand{\arraystretch}{1.12}
\setlength{\tabcolsep}{4pt}
\begin{tabular*}{\avmaintablewidth}{@{\extracolsep{\fill}}l c c c c c@{}}
\toprule
\multirow{2}{*}{\textbf{Verifier}} & \multicolumn{4}{c}{\textbf{Accuracy (\%)}} & \multirow{2}{*}{\textbf{Avg.}} \\
\cmidrule(lr){2-5}
 & \textbf{VerifyBench} & \textbf{VerifyBench-Hard} & \textbf{SCI-VerifyBench} & \textbf{VerifierBench} & \\
\avtabgroup{Rule-based verifiers}
Exact Match & 50.55 & 70.76 & 46.68 & 61.52 & 57.38 \\
Math-Verify & 66.95 & 76.00 & 60.32 & 59.92 & 65.80 \\
\avtabgroup{Model-based verifiers}
XVerify-8B-I & 91.95 & 83.10 & 80.16 & 80.80 & 84.00 \\
CompassVerifier-32B & 92.55 & 87.90 & 86.24 & 90.80 & 89.37 \\
\avtabgroup{Scientific and tool-augmented verifiers}
SCI-Verifier-8B & -- & 90.30 & 86.28 & 93.01 & 89.86\textsuperscript{*} \\
CoSineVerifier-Label-32B & 95.70 & 90.00 & 86.40 & -- & 90.70\textsuperscript{*} \\
CoSineVerifier-Tool-4B & 96.60 & 91.90 & 89.70 & -- & 92.73\textsuperscript{*} \\
\avtabgroup{Same GPT-5.4-Mini verifier interface}
Official prompt & 96.30 & 90.29 & 86.76 & 90.77 & 91.03 \\
\method{} Prompt Only & 96.20 & 90.49 & 86.80 & 91.66 & 91.29 \\
\textbf{\method{} Prompt + Code} & \textbf{96.75} & \textbf{92.43} & \textbf{89.92} & \textbf{93.11} & \textbf{93.05} \\
\bottomrule
\end{tabular*}
\caption{Results of reference-based answer verification benchmarks. The local variants use the same GPT-5.4-Mini binary verifier interface; Prompt + Code adds frozen code modules before the same \method{} prompt verifier. Avg. averages reported columns; starred averages use fewer than four benchmarks.}
\label{tab:main_results}
\end{table*}

\subsection{Frozen Inference and Construction Trace}

\method{} freezes the code modules, prompt, parser, and routing order before benchmark evaluation. This subsection defines how direct decisions and the model-based fallback are combined after construction.

\paragraph{Fallback verifier.}

The model-based fallback verifier $\Vfb$ uses the final \method{} sectioned binary prompt $\Pfb$ produced during construction. It handles examples for which no validated code module can make a direct decision. Prompt optimization uses recurring verifier errors, accepted rule cards, and rejected rule feedback to update prompt guidance while preserving the same binary interface as the official benchmark prompts. Each call receives a question, a reference answer, and a candidate answer, then emits \correct{} or \incorrect{}. This design keeps recurring verifier errors that are not safe to encode as deterministic code available as prompt guidance while reserving deterministic promotion for rule cards with support examples.

\paragraph{Inference and routing.}

At inference, activated code modules run first in a fixed priority order set before evaluation. Let $r^\star$ be the first activated code module that returns a direct decision. If such a module outputs \correct{} or \incorrect{}, $\Vfb$ is not invoked, and later modules cannot override the decision. During construction replay, any candidate whose direct decisions conflict with an earlier promoted module on overlapping support or replay examples is rejected unless it abstains on the overlap or the coordinator updates the fixed priority order and replays the full construction pool. Only when all code modules abstain is the \textbf{fallback verifier} $\Vfb$ consulted:

\[
V_{\mathrm{AV}}(x) =
\begin{cases}
\tau_{r^\star}(\pi_{r^\star}(x)), & \text{if } r^\star \text{ exists},\\
\Vfb(x), & \text{otherwise}.
\end{cases}
\]

This design keeps deterministic decisions guarded and validated by replay while maintaining flexibility for complex or ambiguous cases.

The construction process also records a round trace that is evaluated after the runtime state is frozen.

\section{Experiments}
\label{sec:experiments}

We evaluate \method{} under a frozen construction protocol to test whether recurring verifier errors can be converted into reliable verifier inductive biases. Across four reference-based verification benchmarks, we compare with reported rule-based, model-based, scientific, and tool-augmented verifiers, then use prompt controls, construction replay, direct decision audits, and failure analysis to isolate the contribution of accepted code modules.

\subsection{Experimental Setup}
\label{sec:exp_setup}

\paragraph{Construction data.}
We build \method{} from a construction pool of 5,000 examples designed to expose recurring verifier errors before benchmark evaluation. The pool combines verifier examples from VAR and the TIGER-Lab verifier-sft-data dataset, pairs of questions and reference answers from SuperGPQA and WebInstruct, and constructed diagnostic examples, all normalized into the verifier schema $(q,a^\star,\hat a,y)$. We select examples to cover answer objects and recurring error types, including numeric and unit normalization, symbolic or interval comparison, terminal extraction, closed choice alignment, string or sequence verification, and invalid output cases. During construction, this pool provides residuals, support examples, and counterexamples for learning verifier inductive biases, with source counts and decontamination checks reported in Appendix~\ref{app:data}.

\paragraph{Evaluation benchmarks.}
We evaluate the frozen verifier on four reference-based verification benchmarks: VerifyBench and VerifyBench-Hard \citep{yan2025verifybench}, SCI-VerifyBench \citep{zheng2025sciverifier}, and VerifierBench \citep{liu2025compassverifier}. These benchmarks span general answer verification, hard final answer cases, scientific answer checking, and broad LLM output verification.

\paragraph{Baselines.}
We group baselines by verification style. Rule-based verifiers include Exact Match \citep{yan2025verifybench} and Math-Verify \citep{kydlicek2024mathverify}. Model-based verifiers include XVerify \citep{chen2025xverify} and CompassVerifier \citep{liu2025compassverifier}. Scientific and tool-augmented verifiers include SCI-Verifier \citep{zheng2025sciverifier} and CoSineVerifier \citep{feng2025cosineverifier}. For controlled attribution, we also include GPT-5.4-Mini variants under the same binary verifier interface.

\paragraph{Metrics.}
We use accuracy as the primary metric. We report macro accuracy across benchmarks and use micro accuracy weighted by dataset size for the controlled prompt comparison. We also report code module coverage, fallback verifier accuracy on examples for which all code modules abstain, fallback call reduction, and direct decision audits from evaluation records.

\begin{figure*}[t]
\centering
\includegraphics[width=0.95\textwidth]{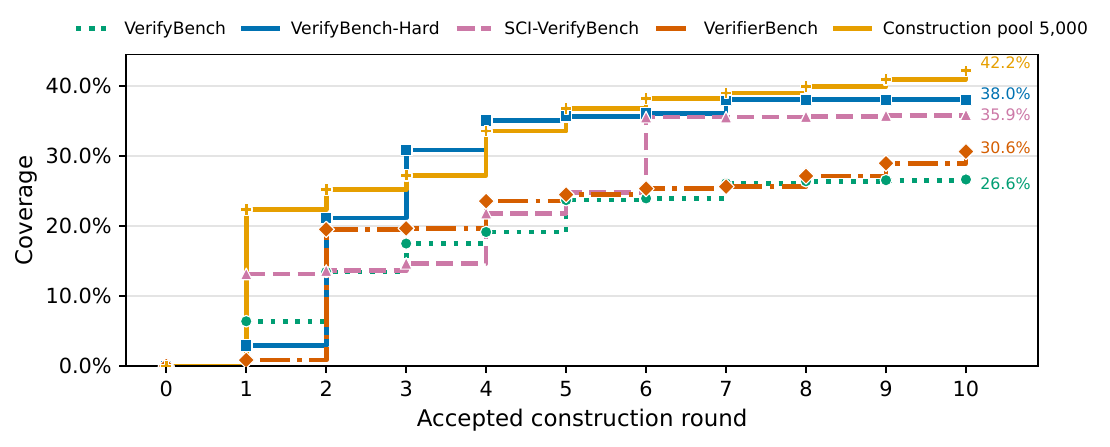}
\caption{Per round replay trace. Round 0 is the baseline without code modules and Round 10 is the frozen verifier. Curves plot cumulative direct coverage on four evaluation benchmarks and the frozen construction monitor of 5,000 examples.}
\label{fig:iteration_trace}
\end{figure*}

\paragraph{Construction and freeze protocol.}
We separate construction from benchmark evaluation by freezing every verifier component before scoring. During construction, we use GPT-5.5 to propose rule cards, code module candidates, and prompt guidance from the construction pool, while GPT-5.4-Mini runs the current fallback verifier during construction replay whenever all active code modules abstain. We admit candidate updates only after structured parsing and replay validation. Before benchmark evaluation, we freeze the module library, fallback prompt, parser, module priority order, and evaluation scripts. During benchmark evaluation, the same GPT-5.4-Mini fallback verifier is called only for examples for which all frozen code modules abstain.
Appendix~\ref{app:construction_block} reports construction data, support statistics, decontamination checks, model configuration, and LLM call accounting.

\subsection{Main Results}
\label{sec:exp_main_results}

\paragraph{Benchmark comparison.}
We demonstrate that \method{} consistently outperforms rule-based, model-based, scientific, and tool-augmented verifiers across four answer-verification benchmarks. Table~\ref{tab:main_results} reports the full accuracy comparison, where \method{} achieves the best result on every benchmark. Relative to the best compared baseline on each benchmark, \method{} improves accuracy by +0.15 on VerifyBench, +0.53 on VerifyBench-Hard, +0.22 on SCI-VerifyBench, and +0.10 on VerifierBench. The final frozen verifier reaches 93.05 macro accuracy across the four benchmarks.

\paragraph{Controlled prompt comparison.}
We attribute most of the controlled gain to adding code modules, whereas prompt rewriting alone gives only a small mixed improvement. All prompt controls use GPT-5.4-Mini and the same binary verifier interface. We replay the public benchmark prompt format in the official prompt setting, and we use the sectioned binary prompt produced during construction in the \method{} prompt setting. For Prompt + Code, we keep the fallback model, final prompt, parser, and binary interface fixed, adding only code modules before the prompt verifier. Prompt rewriting raises macro accuracy from 91.03 to 91.29, while adding code modules raises it to 93.05 with positive gains on every benchmark. The +1.76 macro point gain also holds under micro accuracy weighted by dataset size, where the three local variants score 90.84, 91.15, and 92.95.

\subsection{Construction Replay Trace}
\label{sec:exp_construction_trace}

\paragraph{Round coverage.}
We demonstrate that accepted construction rounds expand supported direct coverage progressively as recurring residuals are converted into validated code modules. We evaluate the verifier state after each accepted round on the construction monitor and, for reference, on the four benchmarks in Figure~\ref{fig:iteration_trace}. We interpret the stepwise increases as evidence that the final verifier is built through accumulated rule promotion rather than a single late update.

\paragraph{Accepted rule groups.}
We analyze the construction trace by accepted rule groups and observe that early rounds account for most covered construction examples while later rounds add smaller, more specialized updates. By Round 10, the accepted construction path covers 2,109 examples in the construction monitor, which contains 5,000 examples. Rounds 1--5 account for 1,840 of these covered examples, and Rounds 6--10 add 269 more. This split across rounds is also reflected in the accepted groups, with extraction and normalization rules entering earlier and sequence, invalid-output, and boundary rules entering later. Appendix Tables~\ref{tab:construction_generalization_audit} and~\ref{tab:rule_revision_case} report the full construction audit and one revision case.

\subsection{Direct Decision Attribution}
\label{sec:exp_direct_attribution}

\paragraph{Direct coverage and corrections.}
We observe that code modules both make direct decisions and correct errors from the prompt-only verifier. Table~\ref{tab:module_decomp} reports 2,665 direct decisions, corresponding to 32.13\% fewer fallback calls, and 149 prompt-only verifier errors corrected by code modules. Direct coverage is substantial across all four benchmarks (26.65\%--38.04\%), while corrections are largest on SCI-VerifyBench (78), where the fallback verifier is weakest on abstained examples.

\begin{table}[!ht]
\centering
\footnotesize
\setlength{\tabcolsep}{2.2pt}
\begin{tabular}{l c c c c c}
\toprule
\textbf{Bench.} & \textbf{Rules} & \textbf{Direct} & \textbf{Direct\%} & \textbf{Corr.} & \textbf{Fallback} \\
\midrule
VB & 29 & 533 & 26.65 & 11 & 95.57 \\
Hard & 31 & 372 & 38.04 & 19 & 87.79 \\
SCI & 41 & 897 & 35.88 & 78 & 84.28 \\
VerifierBench & 24 & 863 & 30.64 & 41 & 90.07 \\
\midrule
All & -- & 2,665 & 32.13 & 149 & 89.61 \\
\bottomrule
\end{tabular}
\caption{Direct decision attribution. Direct\% is the share of examples handled by code modules, equal to fallback call reduction; Corr. counts prompt errors fixed by code modules. Fallback is prompt verifier accuracy on examples for which all code modules abstain.}
\label{tab:module_decomp}
\end{table}

\paragraph{Coverage versus corrections.}
We interpret direct coverage and accuracy contribution as related signals that are not interchangeable. Corrections are larger where the fallback prompt is weaker, while direct coverage also measures routing. We therefore treat fallback call reduction as a consequence of direct routing, not as the main evidence for inductive bias acquisition. The appendix direct decision audit reports zero benchmark label disagreements for direct decisions made by code modules.

\paragraph{Negative controls.}
We test generic shortcut rules as an alternative explanation, but the controls do not reproduce the gains from code modules. A small generic heuristic control covers 502 examples but introduces 15 regressions, and promotion using support alone admits replay and counterexample violations. We report these controls and fallback diagnostics in Appendix~\ref{app:ablations}, and construction call accounting in Appendix~\ref{app:cost_audit}.

\subsection{Bias Categories and Failure Boundaries}
\label{sec:analysis}

\paragraph{Bias categories.}
We find that learned verifier inductive biases concentrate around recurring answer forms and guard conditions tied to implicit annotation assumptions, specifying when common answer forms should be treated as equivalent or nonequivalent.

\paragraph{Accepted rule categories.}
We observe that no single rule group explains the full improvement, but each accepted group contributes a guarded portion of the gain. Table~\ref{tab:module_group_removal} reports direct decisions, corrected prompt-only verifier errors, and macro accuracy drops for each removed group. These removal results indicate that the gains come from reusable comparison logic for answer forms rather than dataset-specific patches.

\begin{table}[!ht]
\centering
\footnotesize
\setlength{\tabcolsep}{2.5pt}
\begin{tabularx}{\columnwidth}{X r r r}
\toprule
\textbf{Removed group} & \textbf{Direct} & \textbf{Corr.} & \textbf{$\Delta$Avg} \\
\midrule
Invalid or conflicting outputs & 207 & 38 & 0.43 \\
Closed choice / verdict mapping & 1,016 & 20 & 0.33 \\
Numeric / unit normalization & 369 & 22 & 0.26 \\
Boundary / format stress rules & 138 & 24 & 0.21 \\
Symbolic math / intervals & 175 & 19 & 0.21 \\
Terminal / object extraction & 540 & 15 & 0.20 \\
String / sequence verification & 220 & 11 & 0.13 \\
\bottomrule
\end{tabularx}
\caption{Group removal under the same prompt verifier. $\Delta$Avg is the macro accuracy drop when direct decisions from the group are routed back to the prompt verifier.}
\label{tab:module_group_removal}
\end{table}

\paragraph{Failure boundaries.}
We characterize the remaining errors as cases that should remain with the fallback verifier. Code modules route deterministic comparison cases directly and abstain on ambiguous or underspecified judgments, keeping rule activation conservative. In a manual analysis of 80 remaining disagreement cases, 53 involve annotation or ambiguity issues and 27 are remaining verifier errors.

\section{Conclusion}
\label{sec:conclusion}

We demonstrate that \method{} converts recurring verifier errors into auditable comparison logic. It learns verifier inductive biases from construction residuals and promotes supported patterns into code modules or prompt guidance. Across four benchmarks, it reaches 93.05\% macro accuracy, improves the prompt-only verifier by +1.76 points, and cuts fallback calls by 32.13\% without changing model parameters.

\section*{Limitations}
\label{sec:limitations}

The current scope is reference-based binary verification with a fixed construction pool and frozen inference interface. Future work should extend the same construction loop driven by recurring verifier errors to richer grading rubrics, multiple reference answers, and domain-specific evaluation standards while preserving support, counterexample, overlap, and replay validation constraints. Future work should also make construction easier to reproduce and maintain with open construction agents, versioned rule libraries, clearer record visualization, and continuous replay tests as new domains or answer formats are added.

\section*{Ethical Considerations}
\method{} judges candidate answers against references and may be used as a reward component in downstream filtering or training pipelines, so incorrect verifier outputs can mislead data selection when references are incomplete or ambiguous. We keep module activation conservative, retain abstention for deterministic rules, and separate benchmark scores from hard case analysis. Construction and benchmark data should be used only under their licenses and permitted use terms; scientific and medical examples are verification cases, not expert advice. If diagnostic cases are released, they should be anonymized and distributed only under the relevant dataset licenses.

\bibliography{custom}

@misc{yan2025verifybench,
  title = {{VerifyBench}: Benchmarking Reference-based Reward Systems for Large Language Models},
  author = {Yan, Yuchen and Jiang, Jin and Ren, Zhenbang and Li, Yijun and Cai, Xudong and Liu, Yang and Xu, Xin and Zhang, Mengdi and Shao, Jian and Shen, Yongliang and Xiao, Jun and Zhuang, Yueting},
  year = {2025},
  eprint = {2505.15801},
  archivePrefix = {arXiv},
  primaryClass = {cs.CL},
  doi = {10.48550/arXiv.2505.15801},
  url = {https://arxiv.org/abs/2505.15801}
}

@misc{chen2025xverify,
  title = {{xVerify}: Efficient Answer Verifier for Reasoning Model Evaluations},
  author = {Chen, Ding and Yu, Qingchen and Wang, Pengyuan and Hu, Mengting and Zhang, Wentao and Wang, Zhengren and Tang, Bo and Xiong, Feiyu and Li, Xinchi and Wang, Chao and Yang, Minchuan and Li, Zhiyu},
  year = {2025},
  eprint = {2504.10481},
  archivePrefix = {arXiv},
  primaryClass = {cs.CL},
  doi = {10.48550/arXiv.2504.10481},
  url = {https://arxiv.org/abs/2504.10481}
}

@misc{liu2025compassverifier,
  title = {{CompassVerifier}: A Unified and Robust Verifier for {LLM}s Evaluation and Outcome Reward},
  author = {Liu, Shudong and Liu, Hongwei and Liu, Junnan and Xiao, Linchen and Gao, Songyang and Lyu, Chengqi and Gu, Yuzhe and Zhang, Wenwei and Wong, Derek F. and Zhang, Songyang and Chen, Kai},
  year = {2025},
  eprint = {2508.03686},
  archivePrefix = {arXiv},
  primaryClass = {cs.CL},
  doi = {10.48550/arXiv.2508.03686},
  url = {https://arxiv.org/abs/2508.03686}
}

@misc{zheng2025sciverifier,
  title = {{SCI-Verifier}: Scientific Verifier with Thinking},
  author = {Zheng, Shenghe and Huang, Chenyu and Yu, Fangchen and Yao, Junchi and Ye, Jingqi and Chen, Tao and Luo, Yun and Ding, Ning and Bai, Lei and Cui, Ganqu and Ye, Peng},
  year = {2025},
  eprint = {2509.24285},
  archivePrefix = {arXiv},
  primaryClass = {cs.CL},
  doi = {10.48550/arXiv.2509.24285},
  url = {https://arxiv.org/abs/2509.24285}
}

@misc{feng2025cosineverifier,
  title = {{CoSineVerifier}: Tool-Augmented Answer Verification for Computation-Oriented Scientific Questions},
  author = {Feng, Ruixiang and An, Zhenwei and Wen, Yuntao and Le, Ran and Jia, Yiming and Yang, Chen and Chen, Zongchao and Chen, Lisi and Gao, Shen and Shang, Shuo and Song, Yang and Zhang, Tao},
  year = {2025},
  eprint = {2512.01224},
  archivePrefix = {arXiv},
  primaryClass = {cs.LG},
  doi = {10.48550/arXiv.2512.01224},
  url = {https://arxiv.org/abs/2512.01224}
}

@misc{chen2025judgelrm,
  title = {{JudgeLRM}: Large Reasoning Models as a Judge},
  author = {Chen, Nuo and Hu, Zhiyuan and Zou, Qingyun and Wu, Jiaying and Wang, Qian and Hooi, Bryan and He, Bingsheng},
  year = {2025},
  eprint = {2504.00050},
  archivePrefix = {arXiv},
  primaryClass = {cs.CL},
  doi = {10.48550/arXiv.2504.00050},
  url = {https://arxiv.org/abs/2504.00050}
}

@misc{liu2025grm,
  title = {Inference-time Scaling for Generalist Reward Modeling},
  author = {Liu, Zijun and Wang, Peiyi and Xu, Runxin and Ma, Shirong and Ruan, Chong and Li, Peng and Liu, Yang and Wu, Yu},
  year = {2025},
  eprint = {2504.02495},
  archivePrefix = {arXiv},
  primaryClass = {cs.CL},
  doi = {10.48550/arXiv.2504.02495},
  url = {https://arxiv.org/abs/2504.02495}
}

@misc{xia2025agentrm,
  title = {{AgentRM}: Enhancing Agent Generalization with Reward Modeling},
  author = {Xia, Yu and Fan, Jingru and Chen, Weize and Yan, Siyu and Cong, Xin and Zhang, Zhong and Lu, Yaxi and Lin, Yankai and Liu, Zhiyuan and Sun, Maosong},
  year = {2025},
  eprint = {2502.18407},
  archivePrefix = {arXiv},
  primaryClass = {cs.CL},
  doi = {10.48550/arXiv.2502.18407},
  url = {https://arxiv.org/abs/2502.18407}
}

@misc{huang2025pitfalls,
  title = {From Accuracy to Robustness: A Study of Rule- and Model-based Verifiers in Mathematical Reasoning},
  author = {Huang, Yuzhen and Zeng, Weihao and Zeng, Xingshan and Zhu, Qi and He, Junxian},
  year = {2025},
  eprint = {2505.22203},
  archivePrefix = {arXiv},
  primaryClass = {cs.CL},
  doi = {10.48550/arXiv.2505.22203},
  url = {https://arxiv.org/abs/2505.22203}
}

@misc{weng2026learning_beyond_gradients,
  title = {Learning Beyond Gradients},
  author = {Weng, Jiayi},
  year = {2026},
  month = may,
  howpublished = {Blog post},
  url = {https://trinkle23897.github.io/learning-beyond-gradients/}
}

@misc{hao2026recreate,
  title = {{ReCreate}: Reasoning and Creating Domain Agents Driven by Experience},
  author = {Hao, Zhezheng and Wang, Hong and Luo, Jian and Zhang, Jianqing and Zhou, Yuyan and Lin, Qiang and Wang, Can and Dong, Hande and Chen, Jiawei},
  year = {2026},
  eprint = {2601.11100},
  archivePrefix = {arXiv},
  primaryClass = {cs.AI},
  doi = {10.48550/arXiv.2601.11100},
  url = {https://arxiv.org/abs/2601.11100}
}

@misc{wang2023voyager,
  title = {Voyager: An Open-Ended Embodied Agent with Large Language Models},
  author = {Wang, Guanzhi and Xie, Yuqi and Jiang, Yunfan and Mandlekar, Ajay and Xiao, Chaowei and Zhu, Yuke and Fan, Linxi and Anandkumar, Anima},
  year = {2023},
  eprint = {2305.16291},
  archivePrefix = {arXiv},
  primaryClass = {cs.AI},
  doi = {10.48550/arXiv.2305.16291},
  url = {https://arxiv.org/abs/2305.16291}
}

@misc{zheng2025skillweaver,
  title = {{SkillWeaver}: Web Agents can Self-Improve by Discovering and Honing Skills},
  author = {Zheng, Boyuan and Fatemi, Michael Y. and Jin, Xiaolong and Wang, Zora Zhiruo and Gandhi, Apurva and Song, Yueqi and Gu, Yu and Srinivasa, Jayanth and Liu, Gaowen and Neubig, Graham and Su, Yu},
  year = {2025},
  eprint = {2504.07079},
  archivePrefix = {arXiv},
  primaryClass = {cs.AI},
  doi = {10.48550/arXiv.2504.07079},
  url = {https://arxiv.org/abs/2504.07079}
}

@misc{mi2026skillpro,
  title = {{Skill-Pro}: Learning Reusable Skills from Experience via {Non-Parametric PPO} for {LLM} Agents},
  author = {Mi, Qirui and Ma, Zhijian and Yang, Mengyue and Li, Haoxuan and Wang, Yisen and Zhang, Haifeng and Wang, Jun},
  year = {2026},
  eprint = {2602.01869},
  archivePrefix = {arXiv},
  primaryClass = {cs.AI},
  doi = {10.48550/arXiv.2602.01869},
  url = {https://arxiv.org/abs/2602.01869}
}

@article{romera2024funsearch,
  title = {Mathematical discoveries from program search with large language models},
  author = {Romera-Paredes, Bernardino and Barekatain, Mohammadamin and Novikov, Alexander and Balog, Matej and Kumar, M. Pawan and Dupont, Emilien and Ruiz, Francisco J. R. and Ellenberg, Jordan S. and Wang, Pengming and Fawzi, Omar and Kohli, Pushmeet and Fawzi, Alhussein},
  journal = {Nature},
  year = {2024},
  volume = {625},
  pages = {468--475},
  doi = {10.1038/s41586-023-06924-6},
  url = {https://www.nature.com/articles/s41586-023-06924-6}
}

@misc{novikov2025alphaevolve,
  title = {{AlphaEvolve}: A coding agent for scientific and algorithmic discovery},
  author = {Novikov, Alexander and Vu, Ng{\^a}n and Eisenberger, Marvin and Dupont, Emilien and Huang, Po-Sen and Wagner, Adam Zsolt and Shirobokov, Sergey and Kozlovskii, Borislav and Ruiz, Francisco J. R. and Mehrabian, Abbas and Kumar, M. Pawan and See, Abigail and Chaudhuri, Swarat and Holland, George and Davies, Alex and Nowozin, Sebastian and Kohli, Pushmeet and Balog, Matej},
  year = {2025},
  eprint = {2506.13131},
  archivePrefix = {arXiv},
  primaryClass = {cs.AI},
  doi = {10.48550/arXiv.2506.13131},
  url = {https://arxiv.org/abs/2506.13131}
}

@misc{kydlicek2024mathverify,
  title = {{Math-Verify}: Math Verification Library},
  author = {Kydl{\'i}{\v{c}}ek, Hynek},
  year = {2025},
  howpublished = {Software library},
  url = {https://github.com/huggingface/Math-Verify}
}

@misc{tigerlab2025verifiersft,
  title = {{TIGER-Lab}/verifier-sft-data},
  author = {{TIGER-Lab}},
  year = {2025},
  howpublished = {Hugging Face dataset},
  url = {https://huggingface.co/datasets/TIGER-Lab/verifier-sft-data}
}

@misc{map2025supergpqa,
  title = {{SuperGPQA}: Scaling {LLM} Evaluation across 285 Graduate Disciplines},
  author = {{M-A-P Team}},
  year = {2025},
  eprint = {2502.14739},
  archivePrefix = {arXiv},
  primaryClass = {cs.CL},
  doi = {10.48550/arXiv.2502.14739},
  url = {https://arxiv.org/abs/2502.14739}
}

@misc{yue2024mammoth2,
  title = {{MAmmoTH2}: Scaling Instructions from the Web},
  author = {Yue, Xiang and Zheng, Tuney and Zhang, Ge and Chen, Wenhu},
  year = {2024},
  eprint = {2405.03548},
  archivePrefix = {arXiv},
  primaryClass = {cs.CL},
  doi = {10.48550/arXiv.2405.03548},
  url = {https://arxiv.org/abs/2405.03548}
}

@misc{guo2025deepseekr1,
  title = {{DeepSeek-R1}: Incentivizing Reasoning Capability in {LLM}s via Reinforcement Learning},
  author = {{DeepSeek-AI}},
  year = {2025},
  eprint = {2501.12948},
  archivePrefix = {arXiv},
  primaryClass = {cs.CL},
  doi = {10.48550/arXiv.2501.12948},
  url = {https://arxiv.org/abs/2501.12948}
}

@misc{yu2025dapo,
  title = {{DAPO}: An Open-Source {LLM} Reinforcement Learning System at Scale},
  author = {Yu, Qiying and Zhang, Zheng and Zhu, Ruofei and Yuan, Yufeng and Zuo, Xiaochen and Yue, Yu and Dai, Weinan and Fan, Tiantian and Liu, Gaohong and Liu, Lingjun and Liu, Xin and Lin, Haibin and Lin, Zhiqi and Ma, Bole and Sheng, Guangming and Tong, Yuxuan and Zhang, Chi and Zhang, Mofan and Zhang, Wang and Zhu, Hang and Zhu, Jinhua and Chen, Jiaze and Chen, Jiangjie and Wang, Chengyi and Yu, Hongli and Song, Yuxuan and Wei, Xiangpeng and Zhou, Hao and Liu, Jingjing and Ma, Wei-Ying and Zhang, Ya-Qin and Yan, Lin and Qiao, Mu and Wu, Yonghui and Wang, Mingxuan},
  year = {2025},
  eprint = {2503.14476},
  archivePrefix = {arXiv},
  primaryClass = {cs.CL},
  doi = {10.48550/arXiv.2503.14476},
  url = {https://arxiv.org/abs/2503.14476}
}

@misc{shao2024deepseekmath,
  title = {{DeepSeekMath}: Pushing the Limits of Mathematical Reasoning in Open Language Models},
  author = {Shao, Zhihong and Wang, Peiyi and Zhu, Qihao and Xu, Runxin and Song, Junxiao and Bi, Xiao and Zhang, Haowei and Zhang, Mingchuan and Li, Y. K. and Wu, Y. and Guo, Daya},
  year = {2024},
  eprint = {2402.03300},
  archivePrefix = {arXiv},
  primaryClass = {cs.CL},
  doi = {10.48550/arXiv.2402.03300},
  url = {https://arxiv.org/abs/2402.03300}
}

@misc{du2026datamaster,
  title = {{DataMaster}: Data-Centric Autonomous {AI} Research},
  author = {Du, Yaxin and Yang, Xiyuan and Zhou, Zhifan and Liu, Wanxu and Lei, Zixing and Chen, Zimeng and Liu, Fenyi and Wu, Haotian and Cai, Yuzhu and Liu, Zexi and Zhu, Xinyu and Wang, WenHao and Zhang, Linfeng and Qian, Chen and Chen, Siheng},
  year = {2026},
  eprint = {2605.10906},
  archivePrefix = {arXiv},
  primaryClass = {cs.LG},
  doi = {10.48550/arXiv.2605.10906},
  url = {https://arxiv.org/abs/2605.10906}
}

@article{baxter2000model,
  title = {A Model of Inductive Bias Learning},
  author = {Baxter, Jonathan},
  journal = {Journal of Artificial Intelligence Research},
  volume = {12},
  pages = {149--198},
  year = {2000},
  doi = {10.1613/jair.731},
  url = {https://doi.org/10.1613/jair.731}
}

\clearpage
\appendix
\section*{Appendix Overview}
The appendix records implementation, audit, and additional evaluation details that support the main text.
\begin{center}
\small
\setlength{\tabcolsep}{3pt}
\begin{tabularx}{\columnwidth}{p{0.12\columnwidth} X}
\toprule
\textbf{Part} & \textbf{Contents} \\
\midrule
A & \textbf{Section~\ref{app:case_study}. Illustrative construction cases and rule card anatomy.} Anonymized support patterns and compact rule card contracts for terminal extraction, unit normalization, verdict guards, and invalid output guards. \\
B & \textbf{Section~\ref{app:construction_block}. Construction implementation and records.} Construction pipeline, model configuration, data pool composition, leakage audit, round trace, rule revision case, support records, and prompt templates. \\
C & \textbf{Section~\ref{app:evaluation_block}. Additional evaluation, controls, and diagnostics.} Statistical uncertainty, direct decision audit, group removal, heuristic control, promotion check ablation, and the Qwen3-4B fallback diagnostic. \\
D & \textbf{Section~\ref{app:audit_block}. Audit, cost, and responsible artifact notes.} Manual error analysis, construction and inference cost accounting, artifact access policy, and screening notes. \\
\bottomrule
\end{tabularx}
\end{center}
\vspace{0.75em}

\section{Illustrative Construction Cases and Rule Card Anatomy}
\label{app:case_study}

This section presents case evidence for construction. It starts from paraphrased construction support examples, then shows the rule card contract that records recurring verifier errors and admits code modules through replay validation. The boxed examples are deliberately compact so that the appendix remains in the ACL two column format.

\avcasebox{Construction Case 1. Unit Implied Numeric Equivalence}{%
\textbf{Training source.} TIGER construction example.

\textbf{Fields.} The question asks for torque in Newton meters. The reference is \texttt{6.79Nm}, while the candidate gives the same terminal value as the bare number \texttt{6.79}.

\textbf{Why the table form was insufficient.} A table entry saying ``numeric/unit normalization'' hides the actual scale choice. The verifier may accept the missing unit only when the task fixes the unit and the candidate supplies a unique terminal scalar.

\textbf{Rule action.} The accepted rule extracts the terminal scalar, checks that the question supplies the omitted unit, and returns \correct{} only for the supported unit preserving comparison. Unsupported units, multiple final numbers, or unparsed expressions trigger \abstain{}.
}

\avcasebox{Construction Case 2. Closed Choice Polarity Conflict}{%
\textbf{Training source.} VAR construction example.

\textbf{Fields.} The task asks for a \texttt{Yes} or \texttt{No} judgment. The reference is \texttt{Yes}, while the candidate's final answer is \texttt{No} after a longer explanation.

\textbf{Why this belongs in a case box.} The error is not visible from aggregate counts. The verifier must align the final answer with the task's closed vocabulary rather than accept explanatory text around an opposite terminal label.

\textbf{Rule action.} The closed choice rule extracts the unique terminal label from both sides and returns \incorrect{} on the polarity conflict. If a response lacks a unique terminal label or mixes equally strong labels, the rule abstains.
}

\avcasebox{Construction Case 3. Short Answer Slot Mismatch}{%
\textbf{Training source.} VAR construction example.

\textbf{Fields.} The task asks for a named creator. The reference is one named person, while the candidate gives a step-by-step lookup response and commits to a different person.

\textbf{Why this case matters.} The candidate contains plausible explanatory text, but the requested output is a compact short answer object. This is a different failure mode from numeric normalization. The verifier must reject process descriptions or wrong objects when the task asks for a compact final object.

\textbf{Rule action.} The short answer guard detects schema, workflow, or process outputs in short answer slots and returns \incorrect{} when the required final object is absent or replaced by a different terminal object. It abstains on genuine alias or synonym questions that require semantic knowledge outside the deterministic guard.
}

\avcasebox{Rule Card Contract Encoded by the Code Module}{%
\textbf{Activation.} Fire only after terminal object extraction identifies a supported answer object, such as a numeric speed ratio or a closed vocabulary verdict.

\textbf{Procedure.} Normalize the extracted objects, preserve task-level constraints such as units or verdict vocabularies, and compare only the bounded final answer region.

\textbf{Abstention condition.} Abstain on missing terminal answers, conflicting final values, unsupported units or vocabularies, and expressions that cannot be parsed into the expected object type.

\textbf{Counterexamples.} Reject or abstain on values outside tolerance, incompatible final objects, omitted final answers, and responses that mix a correct intermediate object with a different final answer.

\textbf{Promotion evidence.} Admit the rule only after support examples trigger the intended path, counterexamples abstain or match construction labels, overlap checks add new coverage, and full construction replay introduces no direct decision regression.
}

\section{Construction Implementation and Records}
\label{app:construction_block}
\suppressfloats[t]

This section records the implementation details needed to audit how recurring verifier errors become verifier behavior. It includes the construction pipeline, construction pool composition, per round growth trace, representative rule card contracts, and construction agent prompt protocols.

\subsection{Construction Implementation Details}
\label{app:method_details}

Table~\ref{tab:method_pipeline} summarizes the construction round pipeline and the audit contract for each stage.

\begin{table*}[t]
\centering
\small
\setlength{\tabcolsep}{3pt}
\begin{tabularx}{\textwidth}{r p{0.22\textwidth} X}
\toprule
\textbf{Step} & \textbf{Operation} & \textbf{Audit contract} \\
\midrule
1 & Normalize construction examples & Convert all sources into $(q,a^\star,\hat a,y)$ with source, domain, answer object, and surface form metadata. \\
2 & Decontaminate & Remove exact and fuzzy question/reference overlaps with all evaluation benchmarks before rule construction. \\
3 & Collect residuals & Run the base verifier in the first round and the current verifier in later rounds; retain label mismatches in the residual buffer. \\
4 & Cluster errors & Group residuals by answer object and observed error type, rather than by benchmark identity or evaluation labels. \\
5 & Draft rule cards & For each cluster, specify activation, deterministic procedure, termination rule, support examples, and counterexamples. \\
6 & Audit code cards & Compare each code card with the current module library; repair or merge an existing code module when it has the right behavior type but incomplete guards, add a new code module only when no existing path covers the behavior, and send cards that require model judgment to prompt guidance. \\
7 & Promote or reject & Accept a code update only if support, counterexamples, overlap checks, and full construction replay satisfy the direct decision promotion contract. \\
8 & Freeze runtime state & Lock module order, prompts, parsers, routing, and output contracts after the construction rounds. \\
\bottomrule
\end{tabularx}
\caption{Construction round pipeline. Each accepted update is generated from construction replay on the construction pool and promoted by replay validation before the runtime is frozen.}
\label{tab:method_pipeline}
\end{table*}

\begin{table*}[t]
\centering
\small
\setlength{\tabcolsep}{3pt}
\begin{tabularx}{\textwidth}{p{0.21\textwidth} p{0.15\textwidth} p{0.23\textwidth} X}
\toprule
\textbf{Component} & \textbf{Model} & \textbf{Reasoning/decoding} & \textbf{Role and evaluation boundary} \\
\midrule
Rule mining agent & GPT-5.5 & xhigh reasoning & Construction only; records recurring verifier error clusters in rule cards. \\
Code generation agent & GPT-5.5 & xhigh reasoning & Construction only; audits code cards against the current library, then proposes repairs, merges, new code modules, and tests. \\
Prompt optimization agent & GPT-5.5 & xhigh reasoning & Construction only; incorporates rule card evidence that requires model judgment into prompt guidance for the sectioned binary prompt. \\
Fallback verifier $\Vfb$ & GPT-5.4-Mini & deterministic decoding when supported & Inference/evaluation only on examples for which all code modules abstain; outputs are parsed by a strict binary parser. \\
\bottomrule
\end{tabularx}
\caption{Model configuration for construction and inference. Construction agents operate only on construction replay records and cannot promote code modules without replay validation; the module library, prompt, parser, and routing order are fixed before reported inference runs.}
\label{tab:model_config}
\end{table*}

The implementation follows the pipeline in Table~\ref{tab:method_pipeline}.  It initializes an empty module library and a base prompt verifier, then replays the current verifier, clusters recurring verifier errors, drafts rule cards, audits code cards, and promotes only candidates that pass support, counterexample, overlap, and full replay validation.  It repairs or merges existing code modules when a cluster exposes an existing module gap, adds a new code module only for uncovered deterministic behavior, incorporates cards that require model judgment into prompt guidance, and freezes the module order, prompt, parser, routing order, and output contract.  The design adapts recurring verifier errors and procedural memory to non-parametric verifier optimization with replay validation. The empirical claim remains restricted to the frozen reference-based answer verification experiments reported in the paper.

\subsection{Construction Data Details}
\label{app:data}

Construction data is converted to the canonical $(q,a^\star,\hat{a},y)$ schema.  VAR train \citep{chen2025xverify}, the TIGER-Lab verifier-sft-data dataset \citep{tigerlab2025verifiersft}, and SuperGPQA-Records \citep{map2025supergpqa} are natively compatible after field parsing.  WebInstruct-verified \citep{yue2024mammoth2} is used only as a source of questions with reference answers; derived verifier examples are constructed by generating candidate responses and assigning binary labels.  Construction samples are decontaminated against evaluation sources by exact and fuzzy question/reference matching before code module construction.

The construction procedure follows verifier data construction patterns used by recent verifier work, but it is specialized for deterministic rule induction rather than neural fine tuning.  First, all source examples are normalized into a shared verifier schema and assigned source, answer object, and candidate surface metadata.  Second, exact and fuzzy decontamination removes examples overlapping the target evaluation questions or references.  Third, recurring verifier errors from the fixed construction verifier are grouped by observed error type, such as numeric/unit normalization, symbolic equivalence, terminal extraction, closed choice alignment, string/sequence verification, and invalid output detection.  Fourth, candidate rule cards are promoted only when they have positive support examples, counterexamples, and replay validation with no direct decision regression.  We use a single dataset size convention throughout the paper, with the construction pool defined as 5,000 source-derived and constructed examples.  Internal validation records are implementation records for rule promotion, not separate data pool definitions.

\begin{table}[!t]
\centering
\scriptsize
\setlength{\tabcolsep}{2pt}
\begin{tabularx}{\columnwidth}{p{0.21\columnwidth} p{0.28\columnwidth} X}
\toprule
\textbf{Axis} & \textbf{Coverage} & \textbf{Role} \\
\midrule
Domain/source & VAR 1630; TIGER 1196; SuperGPQA derived 1014; WebInstruct derived 953; constructed diagnostic examples 207 & Supplies construction questions and references across math, science, engineering, medicine, general reasoning, instruction following, short answer verification, closed choice verification, and controlled comparison cases. \\
Answer object & symbolic 1481; numeric 1373; closed choice 1184; short answer 858; open text 77; scientific identifier 13; list/sequence/word set/equation 14 & Covers the object types and comparison surfaces that deterministic verifiers must parse before comparing references and candidates. \\
Error type & closed choice/verdict alignment; numeric/unit normalization; symbolic/interval comparison; terminal extraction; string/sequence verification; invalid or conflicting outputs & Guides error clustering, support examples, and counterexamples for candidate rule cards. \\
\bottomrule
\end{tabularx}
\caption{Two dimensional construction pool design. The domain and source axis supplies broad task coverage, while the answer object and error type axes ensure that construction examples expose the verifier errors that code modules are meant to solve. Counts are from the selected construction pool of 5,000 source-derived and constructed examples; promotion validation is summarized in the text.}
\label{tab:construction_data_matrix}
\end{table}

The construction pool is therefore not a random sample of available records.  It is a selected construction pool of 5,000 examples that combines source records with constructed diagnostic examples and balances domain coverage with answer object and error type coverage.  For each target error type, the construction pipeline selects positive support examples and counterexamples from the construction pool.  The module library is induced only from recurring verifier errors and support examples; every promoted code module must retain live support under promotion validation.

\begin{figure}[t]
\centering
\includegraphics[width=\linewidth]{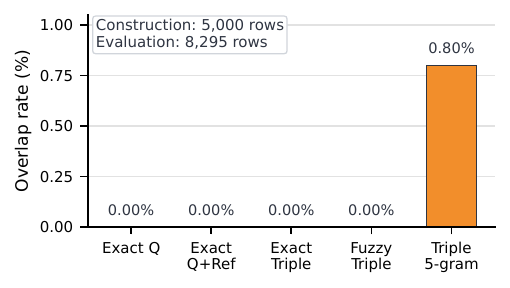}
\caption{Construction--evaluation leakage check for the frozen construction pool. In the contamination-audit format used by data-centric autonomous research systems, the decisive exact and fuzzy verifier-triple checks are zero; the only nonzero plotted signal is low verifier-triple 5-gram surface overlap. Diagnostic question--reference and reference-only repeats are reported in Table~\ref{tab:leakage_audit} because short answers and templates can repeat without copying full verifier examples.}
\label{fig:construction_eval_leakage}
\end{figure}

The final frozen construction pool contains source-derived examples and constructed diagnostic examples.  The constructed examples exercise malformed outputs, invalid-answer conditions, option or verdict conflicts, ordered answer-list surfaces, terminal answer formats, and other answer-format cases that should be covered before deterministic rules are promoted.  Following the leakage-audit practice used in data-centric autonomous research systems, we run a layered construction--evaluation overlap check over exact hashes, fuzzy token overlap, provenance, and 3--5-gram overlap diagnostics \citep{du2026datamaster}.  As shown in Figure~\ref{fig:construction_eval_leakage}, the final construction pool has no exact question, exact question/reference, exact verifier-triple, or fuzzy verifier-triple matches against the 8,295 evaluation examples; the only nonzero plotted signal is low verifier-triple 5-gram surface overlap.  Table~\ref{tab:leakage_audit} reports the corresponding counts and diagnostic-only overlaps.

\begin{table*}[t]
\centering
\small
\setlength{\tabcolsep}{4pt}
\begin{tabularx}{\textwidth}{p{0.24\textwidth} p{0.19\textwidth} r X}
\toprule
\textbf{Audit check} & \textbf{Field / threshold} & \textbf{Count} & \textbf{Interpretation} \\
\midrule
Exact question hash & Normalized question & 0 & No construction example copies an evaluation question. \\
Exact question--reference hash & Normalized question + reference & 0 & No exact evaluation question--answer pair appears in construction. \\
Exact verifier-triple hash & Question + reference + candidate & 0 & No exact evaluation verifier example appears in construction. \\
Fuzzy verifier-triple match & Token Jaccard $\geq 0.90$ & 0 & No near-duplicate verifier example is detected under the strict triple check. \\
Fuzzy question--reference match & Token Jaccard $\geq 0.90$ & 84 & Template-level overlap in task wording; retained as diagnostic because the stricter verifier-triple match is zero. \\
Reference-only exact match & Normalized reference & 64,091 & Diagnostic only: short answers repeat across verifier tasks and are not treated as leakage without question or candidate overlap. \\
Verifier-triple 5-gram overlap & Unique 5-grams & 0.80\% & Low surface overlap diagnostic over full verifier triples. \\
\bottomrule
\end{tabularx}
\caption{Construction--evaluation leakage audit for the final construction pool of 5,000 examples against 8,295 benchmark evaluation examples. Exact checks use normalized hashes; fuzzy checks use token-Jaccard matching at threshold 0.90.}
\label{tab:leakage_audit}
\end{table*}

The runtime support records provide a stricter rule view.  The final support set contains 2,096 live source support examples with 0 direct support errors, plus 13 constructed boundary/format validation records, giving 2,109 supported construction examples in the construction monitor over 5,000 examples.  The final support records validate all 63 delivered code module behaviors with 0 missing support entries and 0 runtime errors.  These support records are construction records for rule promotion, not additional data size conventions.

\subsection{Round Trace Records}
\label{app:round_trace}

Figure~\ref{fig:iteration_trace} is generated from the accepted round records. The construction curve uses the single frozen construction pool of 5,000 source-derived and constructed examples, with 2,096 live support examples from source replay plus 13 constructed boundary/format validation examples, for 2,109 covered examples in the construction monitor over 5,000 examples. After the runtime is frozen, we replay the same final module library on the four benchmarks to report direct routing counts of 533 on VerifyBench, 372 on VerifyBench-Hard, 897 on SCI-VerifyBench, and 863 on VerifierBench. The benchmark counts are post-freeze routing summaries, not construction feedback.

\begin{table*}[t]
\centering
\scriptsize
\setlength{\tabcolsep}{3pt}
\renewcommand{\arraystretch}{1.04}
\begin{tabularx}{\textwidth}{c >{\raggedright\arraybackslash}p{0.28\textwidth} c p{0.18\textwidth} >{\raggedright\arraybackslash}X}
\toprule
\textbf{Round} & \textbf{Generalized construction scope} & \textbf{Contracts / groups} & \textbf{Construction coverage} & \textbf{Audit interpretation} \\
\midrule
0 & No code modules & 0 / 0 & 0 $\rightarrow$ 0 & Baseline with all construction examples routed to the fallback verifier. \\
1 & Exact object and terminal slot rules & 5 / 5 & +1,117 $\rightarrow$ 1,117 & Repeated exact terminal objects are consolidated into reusable activation regions instead of exceptions tied to individual examples. \\
2 & Closed option terminal rules & 12 / 12 & +144 $\rightarrow$ 1,261 & Option rules generalize over recoverable option maps and terminal option surfaces, with nearby option conflicts retained as counterexamples. \\
3 & Closed vocabulary verdict and polarity rules & 5 / 5 & +102 $\rightarrow$ 1,363 & Verdict and polarity phrases are promoted as typed closed vocabulary comparators rather than free form string matches. \\
4 & Numeric, unit, and scientific notation rules & 16 / 16 & +315 $\rightarrow$ 1,678 & Numeric rules add tolerance, unit, and notation boundaries only when construction replay preserves label aligned direct decisions. \\
5 & Symbolic, formula, and interval rules & 13 / 13 & +162 $\rightarrow$ 1,840 & Algebraic and interval rules broaden comparison beyond exact text while keeping endpoint, set, and surface form counterexamples. \\
6 & Sequence, string, and invalid output rules & 11 / 11 & +70 $\rightarrow$ 1,910 & Structured string and invalid answer guards add direct failures for recurring terminal object errors and abstain outside their typed region. \\
7 & Terminal no answer plus angle/ratio support & 8 / 8 & +40 $\rightarrow$ 1,950 & The audit record keeps narrow terminal non answer, angle, and ratio rules after support reselection and construction replay, not benchmark only hits. \\
8 & Unit rate and opaque identifier rules & 4 / 4 & +45 $\rightarrow$ 1,995 & Identifier and SI rate rules are admitted only with whole terminal extraction and unit context guards, turning earlier boundary fixes into reusable rules. \\
9 & Terminal phrase and short phrase expansions & 4 / 3 & +54 $\rightarrow$ 2,049 & Phrase rules add bounded terminal window containment; broader substring candidates remain rejected or routed to fallback. \\
10 & Scientific relative, prefix, unit bare, and boundary/format rules & 5 / 5 & +60 $\rightarrow$ 2,109 & The final checkpoint closes the construction monitor used by the figure; the separate support closure replay validates all 63 delivered code module behaviors. \\
\bottomrule
\end{tabularx}
\caption{Construction generalization audit trace for the ten accepted optimization rounds. Coverage is measured only on the frozen construction monitor used by Figure~\ref{fig:iteration_trace}; benchmark examples are not used to promote a code module. Each row records the generalized rule scope that entered the module library after support, counterexample, overlap, and construction replay validation.}
\label{tab:construction_generalization_audit}
\end{table*}

This table is an audit record rather than a performance curve.  A later round is considered more general when it increases direct coverage on the construction pool through a typed code module behavior that also retains support examples, counterexamples, and a clean construction replay.  Candidates that looked useful but lacked construction support or failed counterexample replay remain in the rejected proposal or support gap records; they are not counted as accepted updates.

\subsection{Construction Side Rule Revision Case}
\label{app:rule_audit_case}

Table~\ref{tab:rule_revision_case} expands one Round 8 identifier revision from the construction records.  The original support example is a SuperGPQA corn genotype verifier pair. The reference answer is \texttt{AaCCRr}, the candidate answer is \texttt{AaCCrr}, and the construction label is \incorrect{}.  The revision fixes an unsafe compact surface match by preserving case for mixed case scientific identifiers, then promotes a fail rule that fires only when a whole terminal identifier can be extracted from both sides.

\par\smallskip\noindent
\refstepcounter{table}\label{tab:rule_revision_case}
\begin{minipage}{0.985\columnwidth}
\centering
\scriptsize
\setlength{\tabcolsep}{3pt}
\renewcommand{\arraystretch}{0.98}
\begin{tabularx}{\linewidth}{>{\raggedright\arraybackslash}p{0.27\linewidth} >{\raggedright\arraybackslash}X}
\toprule
\textbf{Audit item} & \textbf{Construction side evidence} \\
\midrule
Construction example & SuperGPQA support example with reference \texttt{AaCCRr}, candidate \texttt{AaCCrr}, and label \incorrect{}; internal row identifiers are omitted from the paper. \\
Observed pattern & Mixed case identifier strings may differ only by case or terminal slots; lowercase compact normalization is unsafe for genotype, alloy, spectral class, and formula like identifiers. \\
Generalized edit & Add whole terminal mixed case opaque identifier extraction and route different extracted identifiers to a direct fail. The guard abstains if the reference identifier appears anywhere in the candidate. \\
Same group support & Five SuperGPQA examples support the same group, including \texttt{AaCCRr/AaCCrr}, \texttt{Na\_2HXO\_3/Na\_3XO\_3}, \texttt{M2IIIa/M2IIIb}, \texttt{ZCuPb30/ZPbSn10Cu5}, and \texttt{ACBacb/AaBcBC}. \\
Replay result & Runtime replay returns \incorrect{} with reason code \texttt{terminal\_short\_slot\_mismatch}; clean construction replay has 1,980 direct correct decisions, 0 direct wrong decisions, and 0 runtime errors over 4,997 records. \\
\bottomrule
\end{tabularx}
\vspace{0.3ex}
\begin{minipage}{\linewidth}
\fontsize{4.8}{5.0}\selectfont
\begin{verbatim}
def _is_opaque_identifier_token(token: str) -> bool:
    return bool(
        re.fullmatch(r"[A-Za-z][A-Za-z0-9_-]{5,}", token)
        and re.search(r"[A-Z]{2,}", token)
        and re.search(r"[a-z]", token)
    )

    reference_token = _terminal_reference_opaque_identifier(reference)
    if reference_token is None:
        return None
    if reference_token in str(candidate or ""):
        return None
    candidate_obj = _terminal_candidate_opaque_identifier(candidate)
    if candidate_obj is None:
        return None
    candidate_token = str(candidate_obj.get("token") or "")
    if candidate_token == reference_token:
        return None

if opaque_identifier_mismatch_evidence is not None:
    return _build_fail_decision(
        reason_code="opaque_identifier_terminal_mismatch",
        evidence=evidence,
    )
\end{verbatim}
\end{minipage}
\par\smallskip\noindent{\small\textbf{Table~\thetable.} Code module construction audit case for a mixed case identifier revision. The selected source lines are copied from the frozen runtime snapshot and show the activation guard plus the direct fail routing path.}
\end{minipage}
\par\smallskip

\subsection{Module Library Support}
\label{app:modules}
The final module library contains 63 supported code module behaviors with selected pool or fixture based support.  Support is verified by replaying the accepted code modules on support tests outside evaluation. No direct errors are observed, and all 63 target behaviors have at least one live runtime support hit in the final closure pool.  The largest supported groups include compact exact object extraction, closed vocabulary verdict matching, numeric and unit normalization, single choice terminal mismatch detection, interval endpoint mismatch detection, and invalid answer guards.

Each rule card records an activation condition, deterministic execution logic, abstention conditions, support examples, and counterexamples, following the boxed contract in Section~\ref{app:case_study}. Code modules are promoted only when support examples trigger the intended reason, counterexamples abstain or return the construction label as expected, and full construction replay shows no direct error regression. Rejected rule cards are retained as construction feedback when they are too broad, overlap an existing module without coverage gain, or rely on semantic assumptions that cannot be checked deterministically.

\subsection{Prompt Templates}
\label{app:prompts}
The rule mining prompt consumes recurring verifier errors, answer object metadata, support hints, and counterexamples, and must emit structured rule card candidates with activation, procedure, termination, support examples, counterexamples, and rejection risks.  The code generation prompt consumes an accepted rule card plus parser utilities and replay constraints, and must emit the narrowest code module that returns \correct{}, \incorrect{}, or \abstain{}.  The prompt optimization prompt consumes accepted and rejected rule contracts plus prompt verifier errors, then extracts transferable comparison principles into the same sectioned binary prompt used after module abstention.

The \method{} prompt verifier uses \texttt{Role}, \texttt{Rules}, \texttt{Steps}, and \texttt{Output Contract} blocks; it is not an insertion into the official benchmark prompts, which are used only as baselines.  Benchmark rendering receives the question, reference answer, and candidate answer, and it is always evaluated as a binary classifier over \correct{} and \incorrect{}; there is no third prompt label.  Raw construction examples are not inserted into the evaluation prompt.  Output parsing maps explicit correctness labels to the binary interface and treats unparseable outputs as incorrect under reported evaluation.

\section{Additional Evaluation, Controls, and Diagnostics}
\label{app:evaluation_block}

This section keeps only the evaluation views needed for reviewer interpretation. These views are uncertainty estimates, direct decision audit, controls, promotion check ablations, and fallback diagnostics.  Code module coverage is summarized in the main text; longer category breakdowns, direct decision confusion matrices, and fallback error slices are omitted because they restate the same routed examples at finer granularity.

Representative construction support cases are shown in gray boxes in Section~\ref{app:case_study}. The omitted per benchmark category slices, fallback error concentration, and direct decision group tables are diagnostic views of the same routed examples rather than separate evidence needed to understand the main claim.

\subsection{Statistical Uncertainty}
\label{app:uncertainty}

Table~\ref{tab:stat_uncertainty} reports uncertainty for the \method{} prompt-only and prompt-plus-code variants under the same final prompt. Accuracy intervals use Wilson intervals, and $\Delta$ intervals use a paired normal approximation over per example correctness differences. Under this paired comparison, all four code module gains over the prompt component are positive.

\begin{table*}[t]
\centering
\small
\setlength{\tabcolsep}{4pt}
\begin{tabular}{l r r r r r}
\toprule
\textbf{Benchmark} & \textbf{Rows} & \textbf{Prompt} & \textbf{\method{}} & \textbf{$\Delta$} & \textbf{95\% CI for $\Delta$} \\
\midrule
VerifyBench & 2000 & 96.20 [95.27, 96.95] & 96.75 [95.88, 97.44] & +0.55 & [+0.23, +0.87] \\
VerifyBench-Hard & 978 & 90.49 [88.49, 92.17] & 92.43 [90.60, 93.93] & +1.94 & [+1.08, +2.81] \\
SCI-VerifyBench & 2500 & 86.80 [85.42, 88.07] & 89.92 [88.68, 91.04] & +3.12 & [+2.44, +3.80] \\
VerifierBench & 2817 & 91.66 [90.58, 92.62] & 93.11 [92.12, 93.99] & +1.45 & [+1.01, +1.90] \\
\bottomrule
\end{tabular}
\caption{Statistical uncertainty for \method{} prompt and prompt plus code results under the same prompt verifier. Confidence intervals for $\Delta$ use a paired normal approximation over per example correctness differences.}
\label{tab:stat_uncertainty}
\end{table*}

We do not repeat overall F1, balanced accuracy, MCC, direct decision confusion matrices, and group level Wilson intervals because they restate the same paired evaluation records at a finer granularity; the uncertainty summary in Table~\ref{tab:stat_uncertainty} and the direct audit in Table~\ref{tab:direct_audit} are the necessary checks for the appendix.

Table~\ref{tab:direct_audit} reports the direct decision audit behind the final evaluation records. It lists direct coverage, prompt only errors corrected by code modules, direct decision label distribution, and Wilson lower bounds for observed direct accuracy. The per example records contain module predicted \correct{}/\incorrect{} counts, benchmark label counts, and per label precision/recall.

\begin{table*}[t]
\centering
\small
\setlength{\tabcolsep}{3pt}
\begin{tabular}{l r r r r r r r}
\toprule
\textbf{Benchmark} & \textbf{Direct} & \textbf{P-corr.} & \textbf{P-err. fixed} & \textbf{M-err.} & \textbf{Eq.} & \textbf{Not-eq.} & \textbf{Wilson LB} \\
\midrule
VerifyBench & 533 & 522 & 11 & 0 & 353 & 180 & 99.28 \\
VerifyBench-Hard & 372 & 353 & 19 & 0 & 131 & 241 & 98.98 \\
SCI-VerifyBench & 897 & 819 & 78 & 0 & 571 & 326 & 99.57 \\
VerifierBench & 863 & 822 & 41 & 0 & 621 & 242 & 99.56 \\
\midrule
All & 2,665 & 2,516 & 149 & 0 & 1,676 & 989 & 99.86 \\
\bottomrule
\end{tabular}
\caption{Direct decision audit under the fixed final verifier. P-err. fixed counts direct decisions for which the \method{} prompt verifier is wrong and the code module is correct. M-err. counts code module disagreements with the benchmark label. Wilson LB is the lower bound of a two sided 95\% Wilson interval for direct accuracy.}
\label{tab:direct_audit}
\end{table*}

\subsection{Additional Controls and Diagnostics}
\label{app:ablations}
The additional checks target five alternatives a reviewer may consider. The gain might come from prompt rewriting rather than code modules, be reproducible with generic rule heuristics, rely on one dominant module group, depend on weak replay validation, or be tied to the proprietary GPT fallback. The same prompt comparison addresses the first alternative. The \method{} component that uses only the prompt improves the four benchmark average over official GPT-5.4-Mini prompts from 91.03 to 91.29 (+0.26), although individual benchmark deltas are mixed. The deltas are -0.10 on VerifyBench, +0.20 on VerifyBench-Hard, +0.04 on SCI-VerifyBench, and +0.89 on VerifierBench. Adding code modules to the same prompt verifier then improves VerifyBench from 96.20 to 96.75 (+0.55), VerifyBench-Hard from 90.49 to 92.43 (+1.94), SCI-VerifyBench from 86.80 to 89.92 (+3.12), and VerifierBench from 91.66 to 93.11 (+1.45).  The four benchmark average improves from 91.29 to 93.05 (+1.76) after adding code modules.  This paired comparison fixes both the fallback model and the final prompt, so the gain is attributed to code modules rather than to another prompt change.

Table~\ref{tab:module_group_removal} in the main text reports an offline removal ablation under the same prompt verifier. For each group, direct decisions from that group are replaced by the prompt verdict, while all other code module decisions and prompt calls are unchanged. This isolates the marginal contribution of each module group without introducing additional model calls. A group can therefore have high direct coverage but a modest accuracy drop if most of its direct decisions were already correct under the prompt verifier; the corrected count measures the stricter contribution to accuracy. High value code modules are those that remove systematic prompt failures, not necessarily those that fire most often.

The per benchmark removal and cumulative routing views are omitted from the appendix table set because they are deterministic decompositions of the same evaluation records.  Invalid or conflicting output guards matter most on SCI-VerifyBench, closed choice and verdict mapping matter most on VerifyBench-Hard and VerifierBench, and the cumulative route reaches 2,665 direct decisions, 32.13\% fallback call reduction, and 149 prompt errors corrected after all promoted code modules are enabled.

Table~\ref{tab:handwritten_heuristic} adds a small frozen rule control. It uses only generic exact final answer matching, single choice option comparison, closed vocabulary verdict comparison, and exact numeric equality checks, then falls back to the same prompt verifier on all other examples. The control covers fewer examples than code modules learned from recurring verifier errors, has nonzero direct errors, and slightly reduces final micro accuracy relative to the prompt verifier. This negative control is deliberately conservative. It tests whether the improvement can be explained by obvious deterministic shortcuts before crediting the construction process. Its regressions show why \method{} promotes code modules with abstention guards admitted by replay validation instead of always applying broad exact matching heuristics.

\begin{table*}[t]
\centering
\small
\setlength{\tabcolsep}{4pt}
\begin{tabular}{l r r r r r r r}
\toprule
\textbf{Benchmark} & \textbf{Rows} & \textbf{Direct} & \textbf{Direct Corr.} & \textbf{Coverage} & \textbf{Prompt Acc.} & \textbf{Heur.+Prompt Acc.} & \textbf{Reg.} \\
\midrule
VerifyBench & 2,000 & 278 & 96.76 & 13.90 & 96.20 & 95.80 & 9 \\
VerifyBench-Hard & 978 & 224 & 95.98 & 22.90 & 90.49 & 90.18 & 6 \\
SCI-VerifyBench & 2,500 & 0 & -- & 0.00 & 86.80 & 86.80 & 0 \\
VerifierBench & 2,817 & 0 & -- & 0.00 & 91.66 & 91.66 & 0 \\
\midrule
All & 8,295 & 502 & 96.41 & 6.05 & 91.15 & 91.02 & 15 \\
\bottomrule
\end{tabular}
\caption{Frozen rule heuristic control. The control uses only generic exact normalized final answer match, single choice option comparison, closed vocabulary verdict comparison, and exact numeric equality, with the same prompt verifier as fallback. Reg. counts examples where the prompt verifier was correct but the rule direct decision is wrong. This control performs no model calls and does not use generated \method{} code modules.}
\label{tab:handwritten_heuristic}
\end{table*}

Table~\ref{tab:promotion_gate_ablation} reports an archived candidate promotion check ablation. We collect 200 generated candidate code modules from saved construction records, including support checks, counterexample checks, construction replay summaries, protected construction-slice summaries, and candidate check reports. The strict policy reproduces the full promotion contract on these archived candidates. We then weaken one acceptance condition at a time and measure the direct errors or construction violations that would have entered the module library. This experiment audits the promotion checks over generated candidates using saved construction records, rather than reconstructing alternate checks from scratch. The important pattern is that weaker checks accept more candidates but also admit errors. Removing full replay accepts 51 candidates and admits 17 replay errors, while promotion using support alone accepts 95 candidates but admits replay, protected-slice, and counterexample failures. Thus the promotion check ablation supports precision preserving promotion rather than merely showing that the strict policy is conservative.

\begin{table*}[t]
\centering
\scriptsize
\setlength{\tabcolsep}{3pt}
\begin{tabular}{l r r r r r r r}
\toprule
\textbf{Promotion policy} & \textbf{Eval.} & \textbf{Accept} & \textbf{Replay} & \textbf{Replay err.} & \textbf{Slice err.} & \textbf{Slice fail} & \textbf{Ctr-ex} \\
\midrule
Full checks & 200 & 20 & 20 & 0 & 0 & 0 & 0 \\
w/o protected replay & 102 & 28 & 28 & 0 & 9 & 8 & 0 \\
w/o precision regression check & 102 & 23 & 23 & 3 & 0 & 0 & 0 \\
w/o counterexample check & 88 & 14 & 14 & 0 & 0 & 0 & 21 \\
w/o full replay & 186 & 51 & 32 & 17 & 9 & 3 & 0 \\
Support only & 186 & 95 & 55 & 36 & 22 & 5 & 286 \\
\bottomrule
\end{tabular}
\caption{Archived candidate promotion check ablation over saved construction records. Eval. counts candidates evaluable under the weakened policy; Replay counts accepted candidates with saved full replay summaries. Error columns count replay, protected-slice, and counterexample violations admitted by each policy.}
\label{tab:promotion_gate_ablation}
\end{table*}

Table~\ref{tab:qwen4b_fallback} reports a public 4B class fallback diagnostic over the full benchmark suite. We keep the module library, routing records, parser, and direct decisions fixed, replace only the fallback verifier with local Qwen3-4B, and evaluate all 8,295 benchmark examples. The run uses one record per context, strict binary parsing, and counts unparseable fallback outputs as incorrect. This diagnostic checks whether the same code module protocol improves a fallback model outside GPT under fixed examples and parser settings.

\begin{table*}[t]
\centering
\small
\setlength{\tabcolsep}{4pt}
\begin{tabular}{l r r r r r}
\toprule
\textbf{Benchmark} & \textbf{Rows} & \textbf{Qwen3-4B prompt} & \textbf{Direct decisions} & \textbf{Calls$\downarrow$} & \textbf{Prompt+Code} \\
\midrule
VerifyBench & 2,000 & 95.40 & 533 & 26.65 & 96.40 \\
VerifyBench-Hard & 978 & 90.39 & 372 & 38.04 & 92.43 \\
SCI-VerifyBench & 2,500 & 83.64 & 897 & 35.88 & 86.44 \\
VerifierBench & 2,817 & 87.82 & 863 & 30.64 & 91.20 \\
\midrule
Macro avg. & -- & 89.31 & -- & -- & 91.62 \\
All examples & 8,295 & 88.69 & 2,665 & 32.13 & 91.16 \\
\bottomrule
\end{tabular}
\caption{Full Qwen3-4B open fallback diagnostic. The module library, routing records, parser, and direct decisions are fixed; only the fallback verifier is swapped from GPT-5.4-Mini to local Qwen3-4B. Prompt+Code routes direct decisions through the same code modules and sends abstained examples to Qwen3-4B. Calls$\downarrow$ is the percentage of examples that avoid a Qwen fallback call. Macro averages are unweighted over the four benchmarks; All examples are size weighted.}
\label{tab:qwen4b_fallback}
\end{table*}

Group removal, the frozen rule control, the archived candidate promotion check ablation, and the full Qwen3-4B fallback diagnostic are the completed checks used for the current accuracy attribution and robustness diagnosis. The promotion check ablation supports the practical need for protected construction-slice replay, precision checks, counterexamples, and full replay over archived generated candidates. The Qwen3-4B diagnostic shows that the same code modules improve a public 4B fallback from 89.31 to 91.62 macro accuracy while avoiding 32.13\% of fallback calls. We treat this as a diagnostic of replacing the fallback, not as a claim that the rule library is invariant across all fallback models.

The official prompt baseline uses the public benchmark prompt format with GPT-5.4-Mini and the same binary label parser.  Under the same binary evaluation protocol, the official prompt runs reach 96.30\% on VerifyBench full, 90.29\% on VerifyBench-Hard, 86.76\% on SCI-VerifyBench, and 90.77\% on VerifierBench.

\section{Audit and Cost}
\label{app:audit_block}

This section keeps the audit material needed to interpret the reported numbers. It includes the manual error analysis and construction versus inference cost accounting.

\subsection{Manual Error Analysis}
\label{app:audit}
\begin{figure}[t]
\centering
\includegraphics[width=\linewidth]{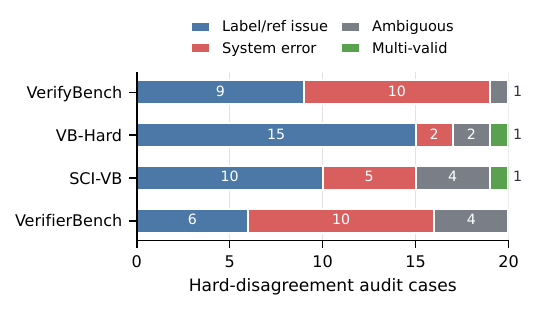}
\caption{Manual error analysis of 80 remaining disagreement cases. Each benchmark contributes 20 cases. Counts are diagnostic only and are not used for construction round rule updates or prompt revision.}
\label{fig:manual_error_analysis}
\end{figure}

We conduct the manual error analysis after removing all direct code module decisions.  For each sampled disagreement, the authors inspect the question, reference answer, candidate answer, benchmark label, and model verdict, without seeing the rule reason or construction history.  Each case is assigned to one of four diagnostic categories: true verifier error, ambiguous or underspecified reference, multiple valid answers, or likely label/reference issue.  An additional author reviews each assignment and records a natural language justification and, for math/science cases, a normalized object comparison when possible.  The analysis covers 80 remaining disagreement cases, 20 from each benchmark, and stores one final diagnostic category per case, summarized in Figure~\ref{fig:manual_error_analysis}.  We use this analysis for hard case diagnosis; benchmark scores continue to use the original benchmark labels, and diagnostic counts are excluded from rule promotion.

The diagnostic categories are true verifier error, ambiguous or underspecified reference, multiple valid answers, and likely label/reference issue.  The accompanying analysis report also records the longer taxonomy of reference sensitive disagreements, cases involving implicit assumptions in annotations, sparse reasoning failures, domain external equivalence, and mixed or overlong responses.  These categories are diagnostic only. Benchmark scores continue to use the original benchmark labels, and diagnostic labels are never fed back into rule promotion.

\subsection{Cost and Audit Accounting}
\label{app:cost_audit}

Table~\ref{tab:cost_audit} separates construction time accounting from inference time savings. Code modules replace fallback calls on direct decisions after the verifier is frozen; construction agent calls are counted separately from reported inference calls.

\begin{table*}[!t]
\centering
\scriptsize
\setlength{\tabcolsep}{3pt}
\begin{tabularx}{\textwidth}{p{0.18\textwidth} p{0.17\textwidth} p{0.13\textwidth} p{0.13\textwidth} X}
\toprule
\textbf{Component} & \textbf{Model/code} & \textbf{Rows/calls} & \textbf{Calls saved} & \textbf{Accounting note} \\
\midrule
Construction agents & GPT-5.5 xhigh & Offline calls & -- & Construction only rule mining, code generation, and prompt optimization.  These calls are counted separately from inference and are not included in fallback-call reduction. \\
Submission session cost & Codex submission sessions & 6.97B tokens & -- & The submission sessions record 6.71B cached input tokens and 18.77M output tokens.  Under standard API token prices, this corresponds to approximately \$765.65 if charged as GPT-5.4-Mini, \$2.55K as GPT-5.4, or \$5.10K as GPT-5.5. \\
Prompt only inference & GPT-5.4-Mini & 8,295 fallback calls & 0 & Controlled baseline with every benchmark example sent to the binary prompt verifier. \\
Prompt + Code inference & GPT-5.4-Mini + code modules & 5,630 fallback calls & 2,665 & Final verifier: code modules route 2,665 examples directly, reducing fallback calls by 32.13\%. \\
Open fallback diagnostic & Qwen3-4B + code modules & 5,630 fallback calls & 2,665 & Same code module records and parser; only the fallback verifier is swapped. \\
Direct code execution & Runtime code & 2,665 direct decisions & -- & Direct decisions have no model token cost at inference. Runtime code, route, rule id, prompt verdict, and final verdict are recorded in the evaluation records. \\
Evaluation audit & Runtime records & All reported examples & -- & Recomputes macro/micro accuracy, direct decision confusion, fallback calls, and paired prompt vs code deltas from the frozen evaluation records. \\
\bottomrule
\end{tabularx}
\caption{Cost and audit accounting. Construction calls are counted separately from inference calls; inference efficiency is reported as fallback call reduction after the verifier is frozen.}
\label{tab:cost_audit}
\end{table*}

\subsection{Responsible Artifact and Checklist Notes}
\label{app:responsible_notes}

Table~\ref{tab:artifact_access} summarizes the artifact access policy used for this work.  We use public or derived sources only for verifier construction, benchmark evaluation, and audit records.  The supplementary package contains an artifact README, requirements file, prompt templates, anonymized synthetic examples, and alignment notes; it should contain derived schemas, hashes, code modules, prompts, routing ledgers, and aggregate audit records rather than redistributing raw third-party examples unless their original licenses and terms permit redistribution.

\begin{table*}[!t]
\centering
\scriptsize
\setlength{\tabcolsep}{3pt}
\begin{tabularx}{\textwidth}{p{0.18\textwidth} p{0.25\textwidth} p{0.20\textwidth} X}
\toprule
\textbf{Artifact class} & \textbf{Examples} & \textbf{Use in this paper} & \textbf{Access and release policy} \\
\midrule
Construction sources & VAR train, TIGER-Lab verifier-sft-data, SuperGPQA-Records, WebInstruct-verified questions & Construction replay, support examples, and counterexamples & Governed by the original source licenses and access terms.  We do not assert a new license over raw source records; derived manifests and hashes can be shared when redistribution of raw examples is not permitted. \\
Constructed diagnostics & Boundary, format, invalid-output, option-conflict, and terminal-answer cases & Construction support and replay validation & Generated verifier examples are released only after removing identifying strings and only with enough metadata to reproduce support and counterexample roles. \\
Evaluation benchmarks & VerifyBench, VerifyBench-Hard, SCI-VerifyBench, and VerifierBench & Frozen benchmark evaluation & Benchmark data remains under the original benchmark licenses and terms.  Evaluation ledgers report benchmark identifiers, routes, module ids, prompt verdicts, final verdicts, parser status, and hashes needed to recompute reported tables. \\
Verifier records & Code modules, prompts, parser settings, priority order, and route ledgers & Reconstructing reported routing, direct decisions, and fallback calls & The release package should exclude API keys, absolute local paths, author-identifying metadata, and non-anonymous repository links.  Model calls to proprietary services are documented by model name and role rather than by undisclosed model parameters. \\
Manual audit records & Author audit of 80 remaining disagreement cases & Diagnostic taxonomy only & We report aggregate categories in the paper.  Case-level diagnostic records should be released only when source licenses allow and after redacting identifying or sensitive text fields. \\
\bottomrule
\end{tabularx}
\caption{Artifact access, intended use, and release policy.  The table distinguishes raw third-party sources from derived records that are sufficient for auditing the reported verifier behavior.}
\label{tab:artifact_access}
\end{table*}

Before sharing derived records, we screen released text fields for personally identifying strings, contact information, author or local path metadata, and offensive content that is not necessary for verifier auditing.  Flagged fields are removed, redacted, or replaced by hashes when the raw text is not needed to recompute a reported result.  The construction and evaluation records are intended for reference-based verifier research and audit replication, not for high-stakes grading, medical advice, or deployment as an expert system.

The manual error analysis in Section~\ref{app:audit} is an internal author audit rather than a recruited human-subjects annotation study.  No external annotators are recruited, no demographic attributes are collected, and the audit labels are not used for rule promotion or benchmark relabeling.  If future work uses external annotators, it should report task instructions, recruitment, compensation, consent, and institutional review or exemption status in the artifact documentation.

\end{document}